\documentclass{article}

\usepackage[preprint]{neurips_2026}

\usepackage[utf8]{inputenc} 
\usepackage[T1]{fontenc}    
\usepackage{hyperref}       
\usepackage{url}            
\usepackage{booktabs}       
\usepackage{amsfonts}       
\usepackage{nicefrac}       
\usepackage{microtype}      
\usepackage{xcolor}         

\usepackage{pifont}
\usepackage{array}           
\usepackage{multirow}          
\usepackage{caption}         
\usepackage{graphicx} 
\usepackage{subcaption}
\usepackage{wrapfig}
\usepackage{enumitem}
\usepackage[table]{xcolor}
\usepackage{colortbl}

\usepackage{amssymb}
\usepackage{tikz}
\usepackage{titletoc}
\usepackage{colortbl}
\usepackage{amsmath}
\usepackage{arydshln}

\definecolor{linkcolor}{rgb}{0.5, 0, 0.7}
\definecolor{citecolor}{rgb}{0.18, 0.57, 0.83}
\hypersetup{
    colorlinks=true,
    linkcolor=blue!60!black,
    citecolor=citecolor,
    filecolor=magenta,      
    urlcolor=magenta,
}

\definecolor{bestcolor}{RGB}{183, 223, 235}     %
\definecolor{secondcolor}{RGB}{225, 242, 248}   %
\definecolor{offlinecolor}{RGB}{232, 245, 233}  %
\definecolor{onlinecolor}{RGB}{255, 243, 224}   %

\newcommand{\best}[1]{\colorbox{bestcolor}{\textbf{#1}}}
\newcommand{\second}[1]{\colorbox{secondcolor}{\underline{#1}}}

\title{Reliability-aware Cross-sample Enhancement for Robust Multimodal Sentiment Analysis}

\author{
Menghua Jiang$^{1}$, 
Haokai Gao$^{2}$, 
Xiangui Kang$^{1}$\thanks{Corresponding authors.},
Haifeng Hu$^{3}$,
Sijie Mai$^{2}$\footnotemark[1]\\
$^1$School of Computer Science and Engineering, Sun Yat-sen University \\
$^2$School of Computer Science, South China Normal University \\
$^3$School of Electronics and Information Technology, Sun Yat-sen University \\
{\tt\small jiangmh26@mail2.sysu.edu.cn}
}

\begin{document}

\maketitle

\begin{abstract}
Multimodal Sentiment Analysis (MSA) aims to infer human emotions from multiple modalities such as text, audio, and vision. In practice, inputs are often corrupted by noise and missing modalities, which degrades performance. Existing methods typically address these challenges in isolation, limiting their effectiveness in realistic settings. To address this limitation, we propose a Reliability-aware Cross-sample Enhancement (RCE) framework. Specifically, RCE first introduces an adaptive variational information bottleneck to model modality-wise uncertainty and perform quality-aware information compression, thereby suppressing redundant noise in unreliable modalities. Furthermore, we design a reliability-aware cross-sample enhancement strategy that retrieves high-confidence, semantically consistent neighbors from a large candidate pool to enrich and calibrate current representations, effectively alleviating information deficiency caused by missing modalities. Building upon this, RCE integrates cross-modal interactions with a multilevel reliability-aware fusion mechanism to adaptively aggregate information across modalities and enhancement stages, leading to more robust multimodal representations. Extensive experiments demonstrate that RCE consistently outperforms state-of-the-art methods across full, noisy, and missing-modality settings.
\end{abstract}

\section{Introduction}
\label{sec:intro}

Multimodal Sentiment Analysis (MSA) integrates text, audio, and vision to better capture complex human emotional states~\cite{CMU-MOSI2016}. With advances in multimodal learning~\cite{EMOE2025,CorrKD2024,PaSE2026}, MSA models leverage complementary cross-modal information, narrowing the gap between human expression and machine understanding, and showing promise in applications such as mental health analysis~\cite{DepMamba2025} and human–robot interaction~\cite{liu2017facial}. However, practical deployment remains challenging: multimodal data are often incomplete, noisy, or uncertain due to imperfect collection~\cite{LNLN2024}, sensor noise~\cite{TMDC2026}, and privacy constraints~\cite{jaiswal2020privacy}, which significantly degrades model reliability and predictive performance.

To achieve optimal multimodal performance, existing studies are often conducted under idealized or isolated settings, either implicitly assuming that all modalities available during training remain fully accessible at inference time, or treating modality missingness and noise contamination as separate problems. Under the full-modality setting, some approaches focus on modeling cross-modal discrepancies. For example, MODS~\cite{MODS2026} captures modality importance via dynamic dominant-modality selection, while DecAlign~\cite{DecAlign2026} achieves cross-modal alignment through prototype-guided optimal transport and distribution matching. To handle noisy inputs, several methods adopt an information-theoretic perspective, where MIB~\cite{MIB2023} introduces an information bottleneck~\cite{tishby2000information} to filter redundant information, while OMIB~\cite{OMIB2025} imposes theoretical constraints on the regularization weights to enable more stable bottleneck representation learning. To address modality missingness, existing approaches typically rely on reconstruction or cross-sample information augmentation. For instance, CyIN~\cite{CyIN2025} reconstructs missing information via cross-modal cycle translation, while HME~\cite{HME2025} enhances representations by retrieving semantically relevant information from other samples.

However, in real-world scenarios, modality missingness and noise are often unknown and dynamically evolving, posing significant challenges to methods developed under idealized or isolated settings~\cite{TMDC2026}. Furthermore, many existing approaches improve robustness to incomplete inputs at the expense of performance in full-modality settings, making it difficult to achieve consistent performance across diverse conditions within a unified framework~\cite{CyIN2025}. Finally, noisy data typically introduce aleatoric uncertainty~\cite{PML2025}, yet most existing denoising strategies overlook this factor and fail to explicitly model modality reliability during the denoising process.

To address the above challenges, we propose a unified framework called Reliability-aware Cross-sample Enhancement (RCE), which learns robust multimodal representations under complex scenarios where noise corruption and modality missingness coexist. RCE consists of three key components: \ding{182} \emph{Adaptive Variational Information Bottleneck}, which maps each modality into a von Mises--Fisher (vMF) distribution parameterized by directional and concentration variables, explicitly capturing modality-wise uncertainty and enabling quality-aware information compression; \ding{183} \emph{Reliability-aware Modality Enhancement}, which mitigates information deficiency caused by missing or degraded modalities by retrieving semantically consistent and high-confidence neighbors from a large candidate pool to enrich and calibrate modality representations; and \ding{184} \emph{Hyper-Modality Generation and Multilevel Fusion}, which models cross-modal interactions via hyper-modality representation learning, integrates enhanced modality features with global contextual information, and employs a multilevel reliability-aware fusion mechanism to adaptively assign weights based on modality confidence, enabling unified modeling from intra-modality to cross-modality interactions. Through these designs, RCE achieves consistent and robust performance across diverse scenarios, including full-modality, noisy, and missing-modality settings. 

Our main contributions are summarized as follows:
\begin{itemize}
    \item We propose a unified framework, termed RCE, to address multimodal sentiment analysis under realistic scenarios encompassing full-modality, noisy, and missing-modality conditions, without requiring separate treatment of each condition.
    
    \item We develop a reliability-aware paradigm that incorporates explicit uncertainty modeling, facilitating adaptive information compression, cross-sample information augmentation, and multi-level fusion for learning robust multimodal representations.
    
    \item Extensive experiments on four benchmark datasets and three evaluation settings demonstrate that RCE consistently outperforms existing methods with stronger robustness.
\end{itemize}

\section{Related Works}

\subsection{Multimodal Sentiment Analysis}
Multimodal Sentiment Analysis (MSA) aims to integrate text, audio, and visual signals to predict human emotional states~\cite{CMU-MOSI2016}. Most existing methods assume that all modalities are available at inference time and focus on designing sophisticated fusion strategies to learn discriminative multimodal representations~\cite{TFN2017,ALMT2023,DEVA2025,MOAC2025,EMOE2025,DLF2025,MMCI2025,CaMIB2025,CaReFlow2026}. Some studies focus on cross-modal alignment, such as DecAlign~\cite{DecAlign2026}, which decouples shared and modality-specific features and leverages prototype-guided optimal transport. Other methods model modality importance, such as MODS~\cite{MODS2026}, which enhances cross-modal interactions through dynamic primary modality selection. In addition, some approaches improve representation capacity by incorporating additional semantic information. For example, PSA-MF~\cite{PSA-MF2026} introduces personality-aware embeddings for fine-grained fusion. Although these methods perform well under full-modality settings, their performance degrades significantly in real-world scenarios with missing modalities, thereby limiting their practical applicability.

\subsection{Multimodal Sentiment Analysis with Missing Modalities}
To address missing modalities, existing MSA methods attempt to compensate for incomplete data from multiple perspectives. One line of work focuses on reconstructing missing modalities from observed ones~\cite{MCTN2019,MMIN2021,DiCMoR2023,GCNet2023,LNLN2024}. For example, IMDer~\cite{IMDer2023} employs a diffusion-based model for reconstruction, while CyIN~\cite{CyIN2025} restores missing information in the latent space via cross-modal cyclic translation. Another line of work adopts a teacher--student framework~\cite{CMAD2025}. In this paradigm, methods such as HRLF~\cite{HRLF2024} leverage a teacher model trained on fully observed multimodal data to guide student models in handling missing modalities through knowledge distillation. In addition, some approaches enhance representations using cross-sample information. CorrKD~\cite{CorrKD2024} proposes a sample-level contrastive distillation mechanism to capture global cross-sample correlations, whereas HME~\cite{HME2025} enriches observed modalities by retrieving semantically relevant clues from other samples, thereby avoiding explicit reconstruction. 

Despite their effectiveness, these methods typically assume noise-free data, which limits their applicability in real-world scenarios with substantial noise. Moreover, HME~\cite{HME2025} and CorrKD~\cite{CorrKD2024} rely on samples within the current mini-batch, leading to unstable and limited neighborhood information. Semantically similar samples with opposite sentiment polarity may further introduce “negative enhancement,” degrading model performance. In contrast, RCE retrieves semantically consistent and high-confidence neighbor representations from a large-scale candidate pool, resulting in more robust multimodal representations. We also provide a review of related work on multimodal sentiment analysis with noisy modalities. Please refer to Appendix~\ref{appendix:related_work_with_noise}.

\begin{figure*}[t]
    \centering
    \includegraphics[width=\linewidth]{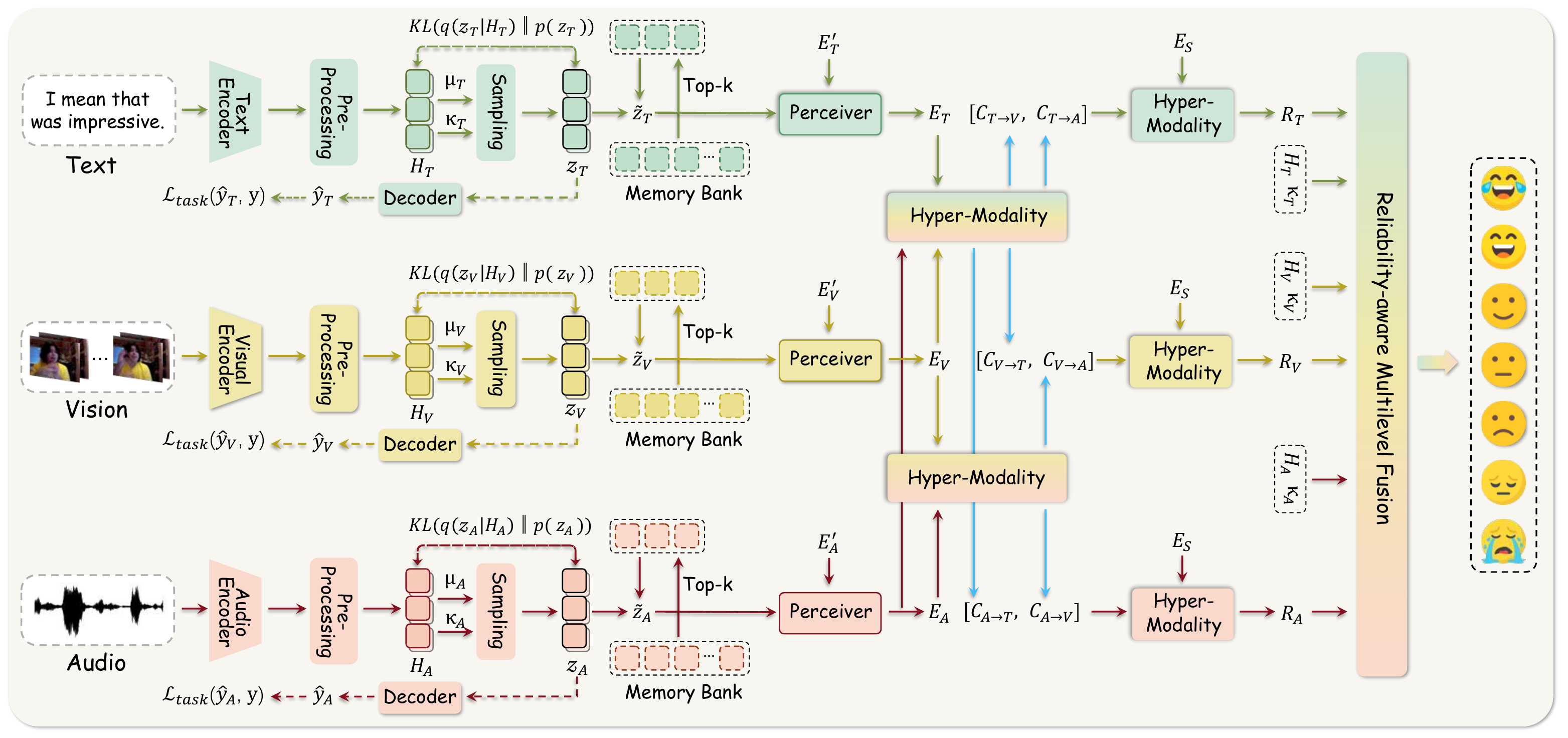}
    \caption{Overview of the RCE framework. For clarity, we provide detailed descriptions and definitions of all notations used in the figure in Appendix~\ref{appendix:notations}.}
    \label{fig:main}
\end{figure*}

\section{Methodology}
\label{sections: methodology}
The overall architecture of the proposed RCE framework is shown in Figure~\ref{fig:main}. It comprises four modules. The Pre-Processing module handles multimodal inputs and extracts modality-specific representations (see Section~\ref{sections_methodology: Pre-Processing Module}). The Adaptive Variational Information Bottleneck (AVIB) module performs reliability-aware adaptive compression (see Section~\ref{sections_methodology: Adaptive Variational Information Bottleneck}). The Reliability-aware Modality Enhancement (RME) module leverages reliable samples to enhance representations (see Section~\ref{sections_methodology: Reliability-aware Modality Enhancement}). Finally, the Hyper-Modality Generation and Multilevel Fusion module generates hyper-modality representations and fuses them with original representations via reliability-aware multilevel fusion (see Section~\ref{sections_methodology: Hyper-Modality Generation and Multilevel Fusion}).

\subsection{Task Formulation}
MSA aims to infer the speaker’s emotional state from video segments comprising textual ($X_T$), acoustic ($X_A$), and visual ($X_V$) modalities. Each modality is represented as a sequence of features, i.e., $X_m \in \mathbb{R}^{L_m \times d_m}$, where $m \in \{T, A, V\}$ indexes the modality. Here, $L_m$ and $d_m$ denote the sequence length and feature dimension of modality $m$, respectively. 
Given a training dataset $\mathcal{D} = \{(X_T^i, X_A^i, X_V^i, y^i)\}_{i=1}^{N}$, the objective is to learn a model $\mathcal{M}_\phi$ that predicts $\hat{y}^i = \mathcal{M}_\phi(X_T^i, X_A^i, X_V^i)$, where $\phi$ denotes the learnable parameters. Depending on the task setting, the prediction $\hat{y}^i$ can be either a continuous sentiment score (regression) or a discrete emotion category (classification).

\subsection{Pre-Processing}
\label{sections_methodology: Pre-Processing Module}
To map heterogeneous multimodal input features into a unified representation space, following prior work~\cite{MulT2019,CorrKD2024,HME2025}, we employ one-dimensional convolutions to project features from each modality, yielding $\hat{X}_m^i = W_{1D}^m(X_m^i)$, where $W_{1D}^m$ denotes the modality-specific one-dimensional convolutional projection layer and $\hat{X}_m^i \in \mathbb{R}^{L_m \times d}$, where $d$ is the shared embedding dimension across modalities. Subsequently, the projected feature sequences are fed into a Transformer~\cite{Transformer2017} encoder to model intra-modal temporal dependencies. Positional encoding is added to preserve sequential order information, yielding modality-specific representations $S_m^i = F_{\phi_m}(\hat{X}_m^i + PE(L_m, d))$, where $F_{\phi_m}$ denotes the modality-specific Transformer encoder with parameters $\phi_m$, and $PE(L_m, d)$ denotes positional encoding of length $L_m$ and dimension $d$. Finally, global max pooling is applied to obtain the initial global feature representation of each modality, i.e., $H_m^i = GMP(S_m^i)$, where $H_m^i \in \mathbb{R}^{d}$ and $GMP(\cdot)$ denotes the global max pooling operation.

\subsection{Adaptive Variational Information Bottleneck}
\label{sections_methodology: Adaptive Variational Information Bottleneck}
To achieve adaptive compression of information conditioned on modality quality, we propose the AVIB module, consisting of two key components: vMF-based uncertainty estimation and quality-aware adaptive compression. We describe each component in detail below.

\paragraph{vMF-based Uncertainty Estimation.} 
To explicitly model sample-level reliability across modalities, we map each modality's feature representation to the von Mises--Fisher (vMF) distribution on the unit hypersphere. For the $i$-th sample and modality $m$, $H_m^i$ parameterizes a posterior distribution over $z$, i.e., $z_m^i \sim \mathrm{vMF}(\mu_m^i, \kappa_m^i)$, where $\mu_m^i$ and $\kappa_m^i$ denote the mean direction and concentration parameter, respectively. Specifically, we define:
\begin{equation}
\mu_m^i = \frac{f_m(H_m^i)}{\|f_m(H_m^i)\|_2} \in \mathbb{S}^{d-1}, \quad
\kappa_m^i = \mathrm{softplus}(g_m(H_m^i)) + \epsilon \in \mathbb{R}^{+},
\label{eq:vmf_params}
\end{equation}
where $f_m(\cdot)$ and $g_m(\cdot)$ are two separate subnetworks used to estimate the directional and concentration parameters, respectively, and $\epsilon$ is a small constant for numerical stability. Under this formulation, $\mu_m^i$ represents the principal direction of the latent representation on the unit hypersphere, while $\kappa_m^i$ characterizes the degree of concentration of the distribution around this direction. A larger $\kappa_m^i$ corresponds to a more concentrated posterior distribution. Following prior work~\cite{CIDER2023,ProCo2024,PML2025}, we interpret it as a proxy for sample-level reliability of the corresponding modality, reflecting how trustworthy the modality is for prediction. Additional details are provided in Appendix~\ref{appendix:von Mises--Fisher_Distribution}.

\paragraph{Quality-aware Adaptive Compression.}
Building on the vMF representation, we model directional information and sample-wise posterior uncertainty within a unified hyperspherical space, enabling modality-quality-aware adaptive compression.

We define the variational posterior as $q(z_m^i \mid H_m^i) = \mathrm{vMF}(\mu_m^i, \kappa_m^i)$, where $\mu_m^i$ and $\kappa_m^i$ are estimated from modality-specific features $H_m^i$. Our goal is to learn a compressed latent representation $z_m^i$ that preserves task-relevant information while discarding redundant and noisy components in the original representation. The corresponding information bottleneck objective is:
\begin{equation}
\mathcal{L}_{\text{VIB}}^m = \beta I(H_m^i; z_m^i) - I(z_m^i; Y),
\end{equation}
where $\beta$ is a hyperparameter controlling the trade-off between compression and predictive power.

Since the mutual information $I(H_m^i; z_m^i)$ is intractable, we use variational inference to approximate it. In the absence of additional prior knowledge and to preserve directional isotropy, we adopt a uniform distribution on the unit hypersphere as the prior: $p(z_m^i) = \mathrm{Uniform}(\mathbb{S}^{d-1})$. Accordingly, the original objective can be reformulated as the following tractable variational objective:
\begin{equation}
\mathcal{L}_{\text{AVIB}}^m = \beta \cdot \mathrm{KL}\Big(q(z_m^i \mid H_m^i)\;\|\;p(z_m^i)\Big) 
- \mathbb{E}_{z_m^i \sim q} \big[\log p(Y \mid z_m^i)\big],
\end{equation}
where the expectation term $\mathbb{E}_{z_m^i \sim q} \big[\log p(Y \mid z_m^i)\big]$ can be implemented via the task loss $\mathcal{L}_{\text{task}}(\hat{y}_m, y)$.

To enable differentiable sampling on the unit hypersphere, we adopt an approximate sampling strategy based on direction--orthogonal decomposition. Given vMF parameters $(\mu, \kappa)$, the latent variable is constructed as:
\begin{equation}
z = w\mu + \sqrt{1 - w^2}\, v,
\end{equation}
where $w \in [-1, 1]$ is a scalar random variable governed by $\kappa$, and $v \sim \mathrm{Uniform}(\mathbb{S}^{d-2})$ is a unit vector sampled from the subspace orthogonal to $\mu$. This construction ensures that $z$ lies on the unit hypersphere $\mathbb{S}^{d-1}$. Under this formulation, the KL divergence admits a closed-form expression:
\begin{equation}
\mathrm{KL}(\mathrm{vMF}(\mu, \kappa) \;\|\; \mathrm{Uniform})
= \kappa \frac{I_{d/2}(\kappa)}{I_{d/2 - 1}(\kappa)} 
+ \log C_d(\kappa) - \log C_d(0),
\end{equation}
when $\kappa = 0$, the vMF distribution reduces to the uniform distribution on the unit hypersphere, with normalization constant $C_d(0) = \frac{\Gamma(d/2)}{2\pi^{d/2}}$.

During training, we introduce auxiliary prediction branches based on modality-specific latent samples and jointly optimize the task loss with the KL regularization. This mechanism assigns a higher information cost to highly concentrated posteriors, encouraging the model to retain them only when the modality provides discriminative information. Conversely, low-quality or noisy modalities are driven toward more dispersed posteriors approaching the uniform prior, reducing their impact on downstream predictions. In this way, AVIB achieves adaptive information compression conditioned on modality quality.


\subsection{Reliability-aware Modality Enhancement}
\label{sections_methodology: Reliability-aware Modality Enhancement}
Inspired by HME~\cite{HME2025}, we enrich each modality by retrieving semantically relevant cues from other samples. However, HME is prone to “negative enhancement” and relies solely on mini-batch samples, leading to insufficient and unstable neighborhood information. To address these limitations, we propose the RME module. Unlike methods that rely only on mini-batch samples, RME maintains a modality-specific memory bank of historical samples, enabling the retrieval of more informative and semantically similar neighbors from a larger candidate pool.

Specifically, for modality $m$, let $z_m^i \in \mathbb{R}^d$ denote the latent representation of the $i$-th sample in the current mini-batch obtained from the AVIB module, with corresponding vMF mean direction $\mu_m^i$ and concentration parameter $\kappa_m^i$. The batch-level representations are denoted as $Z_m^{(b)}=\{z_m^1,\dots,z_m^B\}$, $M_m^{(b)}=\{\mu_m^1,\dots,\mu_m^B\}$, and $K_m^{(b)}=\{\kappa_m^1,\dots,\kappa_m^B\}$, where $B$ is the batch size. We maintain a modality-specific memory bank with capacity $N_b$: $\mathcal{B}_m=\{(z_m^r,\mu_m^r,\kappa_m^r,y^r)\}_{r=1}^{N_m}$ with $N_m \leq N_b$, where $y^r$ is the label. For a given sample $i$, its candidate neighbor set is defined as:
\begin{equation}
\mathcal{C}_i^m =
\begin{cases}
\mathcal{B}_m, & \text{if } N_m = N_b,\\
\{(z_m^j,\mu_m^j,\kappa_m^j,y^j)\mid j \neq i\}, & otherwise.
\end{cases}
\end{equation}
In practice, the memory bank is rebuilt before each training epoch by a forward pass over the training set without gradient updates. The model retrieves neighbors from the memory bank only when it is fully populated, providing a broader and more stable candidate space; otherwise, it falls back to intra-batch retrieval. This design avoids unreliable neighbors in early training while enabling more effective cross-sample complementarity once sufficient historical samples are accumulated.

Notably, to avoid recency bias and ensure balanced retention under limited capacity, we use reservoir sampling~\cite{vitter1985random} for memory updates, such that after observing $t$ samples, each sample is retained with probability $\min(1,N_b/t)$. More details are provided in Appendix~\ref{appendix: reservoir sampling strategy}.

To improve the stability of neighbor retrieval, we do not compute similarity directly on the sampled representations $z_m^i$, but instead perform matching in a more stable reference direction space. Specifically, for a candidate sample $j$ in $\mathcal{C}_i^m$, the similarity is defined as $S_m(i,j)= {\mu_m^i}^{\top}\mu_m^j$, where $\mu_m^i, \mu_m^j$ are unit vectors on the hypersphere. Thus, this is equivalent to cosine similarity~\cite{CorrKD2024}. This design effectively reduces fluctuations caused by sampling noise, making the similarity measure more robust in reflecting semantic consistency between samples. Based on this, we select only the top-$k$ most similar neighbors to form the neighborhood set $\mathcal{N}_i^m$, i.e.,
\begin{equation}
\mathcal{N}_i^m = \text{Top-}k(\mathcal{C}_i^m),
\end{equation}
where $k = \lfloor \rho |\mathcal{C}_i^m| \rfloor$, and $\rho \in (0,1]$ is a proportion hyperparameter. This design helps filter out weakly related candidates and reduces the introduction of irrelevant noise.

To improve the reliability of neighbor aggregation, we incorporate both label distance and confidence constraints into the similarity weighting. For any neighbor $j \in \mathcal{N}_i^m$, its weight is defined as
\begin{equation}
e_{ij}^{(m)} = S_m(i,j) - \alpha |y_i-y_j| + \log(1+\kappa_m^j),
\end{equation}
where $y_i$ and $y_j$ are the labels of the current sample and its neighbor, respectively, and $\alpha$ controls the penalty for label discrepancy. The similarity term captures semantic proximity, the label distance penalizes mismatched labels, and $\log(1+\kappa_m^j)$ favors high-confidence neighbors. The weights are normalized to obtain aggregation coefficients, and the neighbor-enhanced representation is given by
\begin{equation}
\tilde{z}_m^i=\sum_{j\in\mathcal{N}_i^m} w_{ij}^{(m)} z_m^j,
\quad
w_{ij}^{(m)}=\frac{\exp(e_{ij}^{(m)})}{\sum_{k\in\mathcal{N}_i^m}\exp(e_{ik}^{(m)})}.
\end{equation}
If no valid neighbors are available, we set $\tilde{z}_m^i=z_m^i$. This mechanism promotes aggregation from semantically similar, label-consistent, and high-confidence samples, leading to more robust modality enhancement.

Notably, the RME module is used only during training and is not required at inference time. It can also be applied to incomplete data by retrieving semantically relevant cues from available samples, avoiding explicit modality reconstruction and remaining effective under missing modalities.

\subsection{Hyper-Modality Generation and Multilevel Fusion}
\label{sections_methodology: Hyper-Modality Generation and Multilevel Fusion}
\paragraph{Hyper-Modality Representation Generation.}
Next, following HME~\cite{HME2025}, we employ a Perceiver-style latent Transformer~\cite{ALMT2023,LNLN2024} to encode the modality-enhanced representations from the RME module, aiming to obtain compact and context-aware modality-level representations. Specifically, for each modality $m$, we introduce a set of learnable latent prompts $E_m' \in \mathbb{R}^{l_p \times d}$, where $l_p$ is the prompt length and $d$ is the feature dimension. Taking $E_m'$ as queries and the enhanced representation $\tilde{z}_m^i$ as keys and values, we apply cross-attention:
\begin{equation}
\bar{E}_m^i = ATTN_{\psi_m}(E_m', \tilde{z}_m^i) = \text{Softmax}\left(\frac{E_m' W_m^Q (\tilde{z}_m^i W_m^K)^\top}{\sqrt{d}}\right) \tilde{z}_m^i W_m^V,
\end{equation}
where $ATTN_{\psi_m}(\cdot)$ denotes the cross-attention function parameterized by $\psi_m$, and $W_m^Q, W_m^K, W_m^V \in \mathbb{R}^{d \times d}$ are learnable projection matrices. After several standard Transformer blocks, we apply global average pooling over the latent prompt dimension to obtain the modality-level representation: $E_m^i = GAP(\bar{E}_m^i)$, where $E_m^i \in \mathbb{R}^d$ and $GAP(\cdot)$ denotes the global average pooling operation.

Furthermore, to model shared multimodal context, we introduce a shared Perceiver-style Latent Transformer. Unlike the modality-specific branch, this branch takes the original unimodal representations of the current sample as input. Given $H_T^i, H_A^i, H_V^i$, we construct $H_S^i = [H_T^i; H_A^i; H_V^i] \in \mathbb{R}^{3 \times d}$ and apply the same latent cross-attention mechanism to extract global contextual information, yielding the shared representation: $E_S^i = \text{LatentTransformer}_S(H_S^i)$, $E_S^i \in \mathbb{R}^d$.

Subsequently, we leverage cross-modal complementarity to construct hyper-modality representations. For any $m, m_1, m_2 \in \{T, A, V\}$ with $m \neq m_1$, $m \neq m_2$, and $m_1 \neq m_2$, we first model inter-modal interactions via cross-attention:
\begin{equation}
C_{m \to m_1}^i = ATTN_{\psi_{m,m_1}}(E_m^i, E_{m_1}^i), \quad 
C_{m \to m_2}^i = ATTN_{\psi_{m,m_2}}(E_m^i, E_{m_2}^i).
\end{equation}
Then, using the shared representation $E_S^i$ as the query and the interaction features from other modalities as the keys and values, we derive the hyper-modality representation:
\begin{equation}
R_m^i = ATTN_{\psi_{S,m}}(E_S^i, [C_{m \to m_1}^i; C_{m \to m_2}^i]),
\end{equation}
where $R_m^i \in \mathbb{R}^d$. This allows each modality to absorb complementary information from others while incorporating global shared context, leading to more robust cross-modal representations.

\paragraph{Reliability-aware Multilevel Fusion.}
To integrate multi-granularity information across stages, we design a hierarchical aggregation mechanism from intra-modality fusion to cross-modality weighting.

For each modality $m \in \{T, A, V\}$, we first fuse the original unimodal representation $H_m^i$ with the hyper-modality representation $R_m^i$ under shared context. Specifically, we concatenate them and pass them through a lightweight attention network to obtain fusion logits: $a_m^i = F_m([H_m^i \| R_m^i])$, where $F_m(\cdot)$ denotes a lightweight mapping function composed of a linear layer, Dropout, and ReLU. Since the concentration parameter $\kappa_m^i$ in the vMF distribution reflects representation uncertainty, we use it to adaptively modulate the fusion weights, yielding the intra-modality fused representation:
\begin{equation}
u_m^i = \beta_{m,1}^i H_m^i + \beta_{m,2}^i R_m^i, \quad 
\beta_m^i = \text{Softmax} \left( \frac{a_m^i \odot [1, \kappa_m^i]}{\sqrt{d}} \right),
\end{equation}
where $\odot$ denotes element-wise multiplication and $d$ is the feature dimension. This allows adaptive balancing between unimodal and enhanced representations based on modality reliability.

To obtain a global representation, we further perform reliability-aware fusion over $R_T^i, R_A^i, R_V^i$. We compute fusion logits via concatenation: $a_H^i = F_H([R_T^i \| R_A^i \| R_V^i]) \in \mathbb{R}^3$. The modality reliability weights are defined as $\gamma^i = \text{Softmax}([\kappa_T^i, \kappa_A^i, \kappa_V^i])$. The logits are then modulated and normalized: $\eta^i = \text{Softmax} \left( \frac{a_H^i \odot \gamma^i}{\sqrt{d}} \right)$. The resulting global representation is
\begin{equation}
u_H^i = \eta_T^i R_T^i + \eta_A^i R_A^i + \eta_V^i R_V^i.
\end{equation}

Finally, we form four tokens by combining the three intra-modality representations and the global representation: $U^i = [u_T^i; u_A^i; u_V^i; u_H^i] \in \mathbb{R}^{4 \times d}$, which are fed into a two-layer Transformer encoder to model cross-level interactions: $\hat{U}^i = \text{TransformerEncoder}(U^i)$. The output tokens are concatenated to form the final joint representation: $r^i = [\hat{u}_T^i \| \hat{u}_A^i \| \hat{u}_V^i \| \hat{u}_H^i] \in \mathbb{R}^{4d}$. The final representation $r^i$ is then passed to a two-layer MLP to produce the prediction: $\hat{y}^i = \text{MLP}(r^i)$.

\paragraph{Overall Objective.}
To jointly optimize task performance and latent representation compression, we define the overall objective as
\begin{equation}
\mathcal{L}_{all} = \mathcal{L}_{task}(\hat{y}, y) + \lambda \cdot \mathcal{L}_{\text{AVIB}},
\end{equation}
where $\mathcal{L}_{\text{AVIB}} = \frac{1}{3}\sum_{m \in \{T,A,V\}} \mathcal{L}_{\text{AVIB}}^m$, and $\lambda$ is a hyperparameter that balances the task loss and the information bottleneck regularization. For classification and regression tasks, $\mathcal{L}_{task}$ is defined as
\begin{equation}
\mathcal{L}_{task} =
\begin{cases}
CrossEntropy(\hat{y}, y), & classification \\
|\hat{y} - y|, & regression
\end{cases}.
\end{equation}


\section{Experiments}

\subsection{Experimental Setup}
We evaluate RCE on MSA under full, noisy, and missing-modality settings, where both noisy and missing modalities are evaluated under two protocols. To further assess generalization, we also evaluate it on multimodal humor detection (MHD) and multimodal sarcasm detection (MSD) tasks.

\textbf{Datasets.} 
For the MSA task, we conduct experiments on two widely used datasets, CMU-MOSI~\cite{CMU-MOSI2016} and CMU-MOSEI~\cite{CMU-MOSEI2018}. For the MHD and MSD tasks, we adopt the UR-FUNNY~\cite{UR-FUNNY2019} and MUStARD~\cite{MUStARD2019} datasets, respectively. Details are provided in Appendix~\ref{appendix:datasets_information}.

\textbf{Baselines.} 
We compare RCE with several state-of-the-art methods, including recent full-modality approaches such as DecAlign~\cite{DecAlign2026} \textcolor{gray}{\footnotesize [ICLR'26]}, MODS~\cite{MODS2026} \textcolor{gray}{\footnotesize [AAAI'26]}, and PSA-MF~\cite{PSA-MF2026} \textcolor{gray}{\footnotesize [AAAI'26]}. In addition, we focus on two representative categories of methods that are most relevant to RCE. The first category consists of information bottleneck-based approaches, including C-MIB~\cite{MIB2023} \textcolor{gray}{\footnotesize [TMM'23]}, KAN-MCP~\cite{KAN-MCP2025} \textcolor{gray}{\footnotesize [ACM MM'25]}, OMIB~\cite{OMIB2025} \textcolor{gray}{\footnotesize [ICML'25]}, and CyIN~\cite{CyIN2025} \textcolor{gray}{\footnotesize [NeurIPS'25]}. The second category includes uncertainty-aware fusion methods such as QMF~\cite{QMF2023} \textcolor{gray}{\footnotesize [ICML'23]}, PML~\cite{PML2025} \textcolor{gray}{\footnotesize [IJCAI'25]}, and HME~\cite{HME2025} \textcolor{gray}{\footnotesize [NeurIPS'25]}. Detailed descriptions of these methods are provided in Appendix~\ref{appendix:baselines}.

Due to space limitations, details on feature extraction, implementation, and evaluation metrics are provided in Appendix~\ref{appendix:feature_extraction}, Appendix~\ref{appendix:implementation_details}, and Appendix~\ref{appendix:evaluation_metrics}, respectively. In addition, extensive additional experimental results are provided in Appendix~\ref{appendix:additional_experimental_results} for a more comprehensive analysis.

\subsection{Quantitative Results}

\begin{table}[t]
\centering  
\setlength{\belowcaptionskip}{10pt}
\renewcommand\arraystretch{1.1}
\caption{Comparison on CMU-MOSI and CMU-MOSEI. \setlength{\fboxsep}{0.5pt}\colorbox{bestcolor}{\textbf{Best}} and \setlength{\fboxsep}{0.5pt}\colorbox{secondcolor}{\underline{Second}} results are highlighted.}

\label{tab:mosi_mosei}
\resizebox{\textwidth}{!}{

\begin{tabular}{lccccccccccc}

\toprule[1.5pt]

\multirow{2}{*}{\textbf{Method}} 
& \multicolumn{5}{c}{\textbf{CMU-MOSI}} & & \multicolumn{5}{c}{\textbf{CMU-MOSEI}} \\
\cmidrule{2-6} \cmidrule{8-12}
& \textit{Acc7\%$\uparrow$} & \textit{Acc2\%$\uparrow$} & \textit{F1\%$\uparrow$} & \textit{MAE$\downarrow$} &\textit{Corr$\uparrow$} & & \textit{Acc7\%$\uparrow$} & \textit{Acc2\%$\uparrow$} & \textit{F1\%$\uparrow$} & \textit{MAE$\downarrow$} &\textit{Corr$\uparrow$} \\

\midrule

MODS~\cite{MODS2026}
& \second{49.27}  & 85.83  & 85.96  & 0.688 & 0.798 &
& 54.32  & 85.88  & 86.14  & 0.527  & 0.772 \\ 

PSA-MF~\cite{PSA-MF2026}
& 46.50  & 86.43  & 86.19  & 0.686 & 0.807 &
& 55.00   & 86.30  & 86.28  & 0.521  & 0.774 \\ 

DecAlign~\cite{DecAlign2026}
& 45.07  & 85.75 & 85.82 & 0.735 & 0.811 &
& \second{55.02}  & 86.48 & 86.07 & 0.543 & 0.768 \\

C-MIB~\cite{MIB2023} 
& 47.01  & 87.33 & 87.28 & 0.650 & 0.836 &
& 52.47  & 86.08 & 86.05 & 0.551 & 0.773 \\

ITHP~\cite{ITHP2024} 
& 46.57  & 88.40 & 88.40 & 0.650 & 0.848 &
& 52.44  & 86.80 & 86.83 & 0.529 & 0.789 \\

KAN-MCP~\cite{KAN-MCP2025} 
& 46.86  & 88.85 & \second{88.83} & 0.631 & 0.853 &
& 53.67  & 87.18 & 87.20 & 0.523 & 0.787 \\

OMIB~\cite{OMIB2025} 
& 47.59  & \second{88.85} & 88.79 & 0.636 & 0.844 &
& 53.54  & 87.29 & \second{87.27} & 0.527 & 0.786 \\

QMF~\cite{QMF2023} 
& 47.59  & 88.40 & 88.35 & 0.634 & 0.846 &
& 52.34  & \best{87.35} & 87.21 & 0.530 & 0.789 \\

PML~\cite{PML2025} 
& 47.01  & 88.09 & 88.03 & 0.658 & 0.837 &
& 53.91  & 87.07 & 87.04 & 0.527 & 0.786 \\

HME~\cite{HME2025} 
& 48.32  & 88.09 & 88.11 & \second{0.620} & \second{0.856} &
& 54.25  & 86.82 & 86.89 & \second{0.513} & \second{0.792} \\

\textbf{RCE (Ours) }
& \best{50.22}  & \best{89.31} & \best{89.26} & \best{0.589} & \best{0.861} &
& \best{55.44}  & \second{87.29} & \best{87.33} & \best{0.503} & \best{0.799} \\ 

\bottomrule[1.5pt]

\end{tabular}
}
\end{table}

\textbf{Full-modality Setting.} 
The results under the fully multimodal setting are reported in Table~\ref{tab:mosi_mosei}. This setting is adopted by most existing MSA methods. It can be observed that, compared with state-of-the-art approaches such as DecAlign~\cite{DecAlign2026} and MODS~\cite{MODS2026}, RCE achieves the best performance on nearly all metrics. In particular, RCE significantly outperforms existing methods on \textit{Acc7} and \textit{MAE}, highlighting its capability for fine-grained sentiment prediction. This advantage stems from its reliability-aware compression, cross-sample enhancement, and multilevel fusion, which collectively facilitate the learning of robust and discriminative multimodal representations.

\begin{wrapfigure}{r}{0.48\textwidth} 
    \centering
    \includegraphics[width=0.46\textwidth]{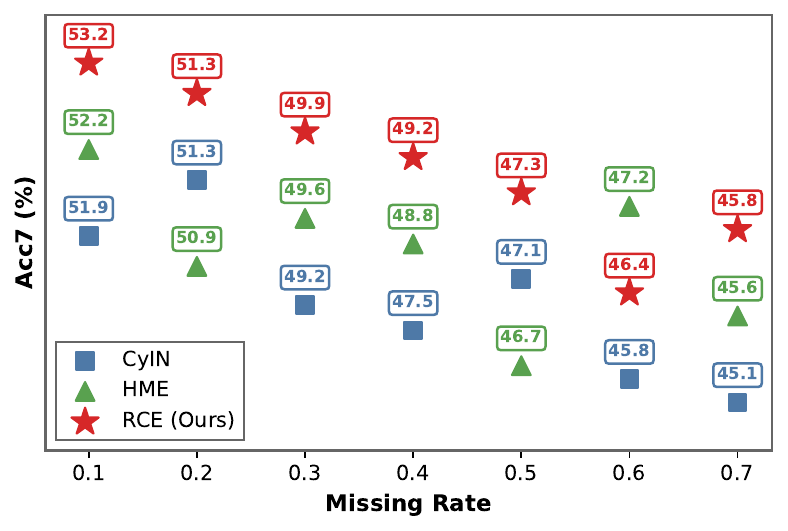}
    \caption{Comparison on CMU-MOSEI under the random missing protocol.}
    \vspace{-2.3em}
    \label{fig:random_missing}
\end{wrapfigure}

\textbf{Missing-modality Setting.} 
We compare RCE with state-of-the-art methods specifically designed for missing modality scenarios, including HME~\cite{HME2025} and CyIN~\cite{CyIN2025}, under the random missing protocol. The results (see Figure~\ref{fig:random_missing}) show that RCE consistently outperforms these methods across most missing rates. Notably, even without modality-specific tuning, RCE demonstrates strong generalization ability. We further provide comparisons under fixed missing protocols against a variety of advanced methods, with results reported in Appendix~\ref{appendix:results_fixed_missing}. Overall, RCE significantly enhances robustness to diverse missing patterns by leveraging informative cues from other reliable and available samples.

\begin{figure}[t]
    \centering
    \begin{minipage}[t]{0.635\linewidth}
        \centering
        \includegraphics[width=\linewidth]{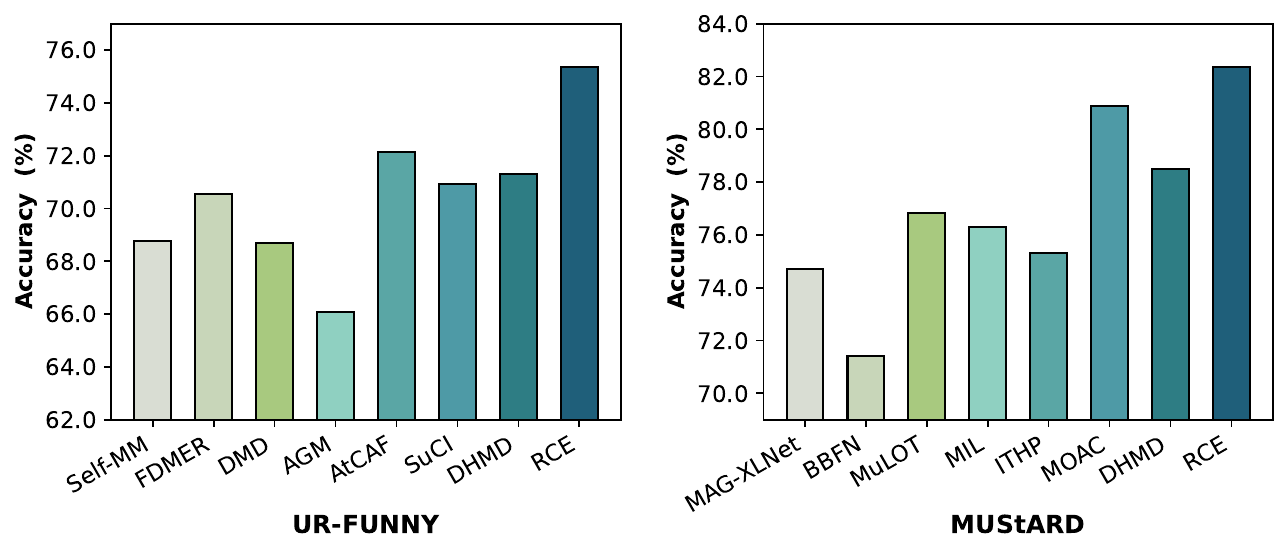}
        \caption{Results on UR-FUNNY and MUStARD.}
        \label{fig:mhd_msd}
    \end{minipage}
    \hfill
    \begin{minipage}[t]{0.345\linewidth}
        \centering
        \includegraphics[width=\linewidth]{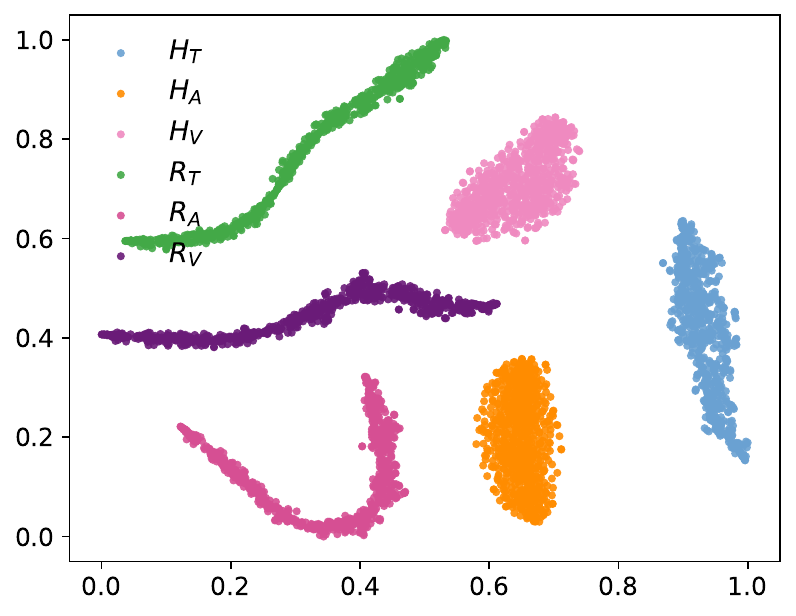}
        \caption{Feature visualization.}
        \label{fig:tsne}
    \end{minipage}
    \vspace{-1em}
\end{figure}

\textbf{MHD and MSD Tasks.}
To further evaluate the generalization capability of RCE on other multimodal tasks, we conduct experiments on the MHD and MSD tasks using the UR-FUNNY~\cite{UR-FUNNY2019} and MUStARD~\cite{MUStARD2019} datasets. As shown in Figure~\ref{fig:mhd_msd}, RCE outperforms the best-performing baselines, AtCAF~\cite{AtCAF2025} and MOAC~\cite{MOAC2025}, by over 3.2\% and 1.5\% in accuracy, respectively, achieving state-of-the-art performance. These results further demonstrate the strong generalization ability of RCE across different multimodal tasks.

\subsection{Ablation Study}
To better understand the contribution of each component in RCE, we conduct an ablation study on several key modules of the proposed framework: (i) `w/o AVIB' removes the AVIB module and directly uses the modality representations without adaptive compression; (ii) `w/o RME' removes the RME module; (iii) `w/o MB' disables the memory bank and only uses mini-batch samples for neighbor retrieval; (iv) `w/o CP' removes the confidence promotion term in neighbor weighting; (v) `w/o HMG' removes the hyper-modality generation component; and (vi) `w/o RF' removes reliability-aware fusion and replaces it with standard attention weighting. The results are reported in Table~\ref{tab:ablation}. Full results and additional variant analyses are provided in Appendix~\ref{appendix:ablation_study}.

\textbf{Effects of the RCE Components.}
Overall, removing most components degrades performance on both CMU-MOSI and CMU-MOSEI, demonstrating the effectiveness of the proposed modules. Among them, `w/o HMG' causes the largest drop on CMU-MOSI, with Acc7 decreasing from 50.22 to 46.72 and MAE increasing from 0.589 to 0.637, highlighting the importance of hyper-modality integration. Removing AVIB also consistently harms performance, indicating that vMF-based uncertainty estimation and adaptive compression effectively suppress noisy modality information. In addition, `w/o MB' yields lower Acc7 on both datasets, suggesting that retrieving reliable neighbors from the memory bank provides more stable enhancement than using only mini-batch samples. Some components exhibit dataset- and metric-dependent effects. Specifically, `w/o CP' causes only minor changes on CMU-MOSEI, while `w/o RF' slightly improves Acc7 but worsens MAE, suggesting that reliability-aware fusion contributes more to prediction calibration than classification accuracy. Although `w/o RME' slightly improves Acc7 on CMU-MOSI, it increases MAE and degrades performance on CMU-MOSEI, indicating that RME generally improves robustness despite minor fluctuations on smaller datasets. Overall, AVIB, MB, and HMG are the most influential components.

\begin{table}[t]
\centering  
\setlength{\belowcaptionskip}{10pt}
\renewcommand\arraystretch{1.1}
\caption{Ablation results of RCE components on CMU-MOSI and CMU-MOSEI. We report \textit{Acc7} and \textit{MAE}. Full results are provided in Appendix~\ref{appendix:ablation_study}.}
\label{tab:ablation}
\resizebox{\textwidth}{!}{
\begin{tabular}{cccccccc}
\toprule[1.5pt]
\textbf{Dataset}
& RCE & w/o AVIB & w/o RME & w/o MB & w/o CP & w/o HMG & w/o RF \\
\midrule

CMU-MOSI
& \second{50.22} / \best{0.589} 
& 48.03 / 0.646 
& \best{50.36} / 0.610 
& 48.03 / 0.616 
& 47.30 / \second{0.610} 
& 46.72 / 0.637 
& 48.47 / 0.626 \\ 

CMU-MOSEI
& \second{55.44} / \best{0.503} 
& 54.60 / 0.518 
& 54.10 / 0.519 
& 53.50 / 0.511 
& 55.14 / \best{0.504} 
& 54.77 / 0.506 
& \best{55.52} / 0.507 \\ 

\bottomrule[1.5pt]
\end{tabular}
}
\vspace{-1em}
\end{table}

\subsection{Further Analysis}

We provide additional analyses in the appendix, including parameter sensitivity analysis (Appendix~\ref{appendix:parameter_sensitivity}), testing stability analysis (Appendix~\ref{appendix:testing_stability}), model complexity analysis (Appendix~\ref{appendix:model_complexity_analysis}), and theoretical complexity analysis (Appendix~\ref{appendix:theoretical_complexity_analysis}).

\textbf{Feature Visualization.}
To investigate the differences between the original modality representations $H_m$ and the generated hyper-modality representations $R_m$, we visualize them using t-SNE~\cite{t-SNE2008}. As shown in Figure~\ref{fig:tsne}, the hyper-modality representations form distinct clusters, indicating that RCE captures complementary sentiment-relevant information beyond unimodal features.

\begin{wraptable}{t}{0.50\textwidth}
\centering
\caption{Training negative enhancement rates on CMU-MOSI and CMU-MOSEI.}
\resizebox{\linewidth}{!}{
\label{tab:negative_enhancement_rate}
\begin{tabular}{ccccc}
\toprule[1.5pt]
\textbf{Dataset} & \textbf{Method} & \textbf{Text} & \textbf{Audio} & \textbf{Vision} \\
\midrule
\multirow{2}{*}{CMU-MOSI}
& HME & 46.29\% & 46.90\% & 46.90\% \\
& RCE & \best{2.60\%} & \best{5.90\%} & \best{6.48\%} \\
\midrule
\multirow{2}{*}{CMU-MOSEI}
& HME & 38.40\% & 38.87\% & 38.87\% \\
& RCE & \best{10.57\%} & \best{31.25\%} & \best{26.08\%} \\
\bottomrule[1.5pt]
\end{tabular}
}\vspace{-1em}
\end{wraptable}

\textbf{Negative Enhancement Analysis.}
As shown in Table~\ref{tab:negative_enhancement_rate}, we report the training-stage negative enhancement rates of HME~\cite{HME2025} and RCE across text, audio, and vision modalities. Compared with HME, RCE consistently reduces negative enhancement cases across all modalities, showing that the proposed reliability-aware and label-aware enhancement strategy effectively suppresses sentiment-conflicting information from neighboring samples. On CMU-MOSI, the rates drop from around 46\% to below 7\%. On CMU-MOSEI, RCE also achieves clear reductions, especially in text, while the smaller improvement in audio may be due to the higher ambiguity of acoustic sentiment cues. Detailed results and analysis are provided in Appendix~\ref{appendix:analysis_negative_enhancement}.




\section{Conclusion}
In this paper, we propose Reliability-aware Cross-sample Enhancement (RCE), a unified framework for robust multimodal sentiment analysis under full, noisy, and missing-modality settings. RCE learns robust multimodal representations by estimating modality reliability with adaptive variational information bottleneck, retrieving reliable cross-sample cues for modality enhancement, and performing reliability-aware multilevel fusion for prediction.

\textbf{Limitations.}
RCE introduces additional hyperparameters and a larger number of parameters. Although it is generally stable within moderate parameter ranges, dataset-specific tuning may still be needed. Future work will explore adaptive and lightweight enhancement strategies.

\bibliography{reference.bib}
\bibliographystyle{plain}




\clearpage

\appendix

\vbox{%
  \hsize\textwidth
  \linewidth\hsize
  \vskip 0.1in
  \hrule height 4pt
  \vskip 0.25in
  \vskip -\parskip
  \centering
  {\LARGE\bf Reliability-aware Cross-sample Enhancement for Robust Multimodal Sentiment Analysis\\[0.15cm](Appendix)}
  \vskip 0.29in
  \vskip -\parskip
  \hrule height 1pt
  \vskip 0.09in
}

\startcontents[appendix]
\vspace{0.5cm}
\printcontents[appendix]{}{1}{\setcounter{tocdepth}{2}}
\vspace{1cm}

\clearpage

\section*{The Use of Large Language Models (LLMs)}
In this work, large language models (LLMs) were used solely for language polishing to improve clarity and fluency. All scientific content, including the experimental design, results, and conclusions, is the responsibility of the authors. No LLM was involved in the ideation, analysis, or interpretation of the research, and all LLM-assisted content was carefully reviewed and verified by the authors.

\section{Notations}
\label{appendix:notations}

We provide a comprehensive overview of the commonly used notations and their definitions in Table~\ref{tab:notation}.

\begin{table}[h]
\centering
\setlength{\belowcaptionskip}{4pt}
\renewcommand\arraystretch{1.1}
\caption{Notation and Definitions}
\label{tab:notation}
\begin{tabular}{ll}
\toprule[1.5pt]
\textbf{Notation} & \textbf{Definition} \\
\midrule

$X_m^i$ & Input feature sequence of modality $m$ for sample $i$.  \\
$L_m$ & Sequence length of modality $m$.  \\
$d_m$ & Original feature dimension of modality $m$.  \\
$d$ & Shared embedding dimension across all modalities.  \\
$\hat{X}_m^i$ & Projected modality feature after 1D convolution.  \\
$S_m^i$ & Sequence representation after modality-specific Transformer encoder.  \\
$H_m^i$ & Modality-specific representation obtained after pre-processing.  \\
$z_m^i$ & Latent representation sampled from AVIB.  \\
$\mu_m^i$ & Mean direction of the vMF distribution.  \\
$\kappa_m^i$ & Concentration parameter of the vMF distribution reflecting modality reliability.  \\
$q(z_m^i \mid H_m^i)$ & Variational posterior of modality $m$.  \\
$p(z_m^i)$ & Uniform prior distribution on the unit hypersphere.  \\
$\mathcal{B}_m$ & Memory bank for modality $m$.  \\
$\mathcal{C}_i^m$ & Candidate neighbor set for sample $i$.  \\
$\mathcal{N}_i^m$ & Selected top-$k$ neighbors.  \\
$S_m(i,j)$ & Directional similarity between sample $i$ and $j$.  \\
$w_{ij}^{(m)}$ & Aggregation weight for neighbor $j$.  \\
$\tilde{z}_m^i$ & Enhanced modality representation from RME.  \\
$\alpha$ & Label discrepancy penalty coefficient.  \\
$\rho$ & Proportion for selecting top-$k$ neighbors.  \\
$l_p$ & Length of latent prompts.  \\
$E_m^i$ & Modality-level latent representation.  \\
$E_S^i$ & Shared cross-modal representation.  \\
$R_m^i$ & Hyper-modality representation.  \\
$a_m^i$ & Fusion logits for intra-modality aggregation.  \\
$\beta_m^i$ & Adaptive fusion weights for intra-modality fusion.  \\
$u_m^i$ & Intra-modality fused representation.  \\
$\gamma^i$ & Reliability weights derived from modality concentration.  \\
$\eta^i$ & Reliability-aware modality fusion weights.  \\
$u_H^i$ & Global multimodal representation.  \\
$U^i$ & Token set combining intra- and inter-modality representations.  \\
$r^i$ & Final joint representation.  \\

\bottomrule[1.5pt]
\end{tabular}
\end{table}

\clearpage

\section{More Related Work}

\subsection{Multimodal Sentiment Analysis with Noisy Modalities}
\label{appendix:related_work_with_noise}
Existing approaches for handling modality noise can be broadly categorized into three groups.
The first group, noise augmentation methods, simulates real-world noise by injecting perturbations into features or raw inputs and developing corresponding defenses. 
For example, Robust-MSA~\cite{Robust-MSA2023} constructs an interactive analysis platform to visualize the impact of modality-specific noise on model performance, and incorporates simple defense strategies to improve interpretability. 
Similarly, MSA-Robustness~\cite{MSA-Robustness2022} introduces a suite of diagnostic tools to assess the sensitivity of multimodal models to noise in individual modalities, and systematically explores robust training strategies to enhance stability under noisy perturbations. 
However, designing dedicated defenses for different noise types is complex and often lacks generalization.

The second group focuses on noise identification and filtering mechanisms to suppress noisy signals and improve the robustness of multimodal representations~\cite{gong2024adaptive}.
For instance, DynMM~\cite{DynMM2023} introduces a dynamic gating mechanism that adaptively adjusts modality contributions or fusion pathways based on input features, thereby mitigating the influence of noisy modalities.
However, such implicit weighting strategies lack explicit reliability modeling and thus remain susceptible to noise.
To further improve modality contribution estimation, Multimodal Boosting~\cite{Multimodal-Boosting2024} leverages unimodal prediction losses as supervisory signals to identify noisy modalities.
Nevertheless, this approach relies on the construction of absolute labels, which are difficult to define precisely and may introduce additional noise during training.

The third group adopts representation regularization strategies, which have emerged as a promising direction. 
These methods typically leverage techniques such as tensor rank minimization~\cite{liang2019learning} or information-theoretic principles~\cite{tishby2000information} to extract discriminative representations from noisy features. 
For example, MIB~\cite{MIB2023}, based on the information bottleneck principle~\cite{tishby2000information}, suppresses redundancy and noise in both unimodal and multimodal representations by learning minimal sufficient task-relevant features. 
ITHP~\cite{ITHP2024} designates a primary modality while treating the remaining modalities as auxiliary detectors to extract salient information. 
OMIB~\cite{OMIB2025} further extends the multimodal information bottleneck framework by constraining regularization weights to ensure an optimal bottleneck representation. 
Similarly, KAN-MCP~\cite{KAN-MCP2025} integrates cross-modal modeling with Dimensionality Reduction and Denoising Modal Information Bottleneck (DRD-MIB)-based feature compression to learn more robust representations.

Despite the success of representation regularization methods, particularly those based on the information bottleneck principle, such approaches typically compress all modalities equally and lack the ability to distinguish modality reliability, which makes them less effective in the presence of noise or heterogeneous modality quality.
In contrast, RCE introduces an Adaptive Variational Information Bottleneck (AVIB) in the von Mises–Fisher (vMF) representation space. Specifically, AVIB jointly models directional information and sample-wise posterior uncertainty within a unified hyperspherical space, enabling reliability-aware adaptive compression that preserves informative signals from reliable modalities while suppressing noise from less reliable ones.
\section{Methodological Details}

\subsection{von Mises--Fisher Distribution}
\label{appendix:von Mises--Fisher_Distribution}
In this section, we provide a detailed description of the von Mises--Fisher (vMF) distribution and its role in modeling sample-level uncertainty in our RCE framework.

\paragraph{vMF-based Uncertainty Estimation.}
To explicitly model sample-level reliability differences across modalities, we map the feature representation of each modality onto a probability distribution defined on the unit hypersphere. Specifically, for the $i$-th sample and modality $m$, the representation $H_m^i$ is used to parameterize a von Mises--Fisher (vMF) posterior distribution:
\begin{equation}
z_m^i \sim \mathcal{V}_d(\mu_m^i, \kappa_m^i),
\end{equation}
where $\mu_m^i \in \mathbb{S}^{d-1}$ denotes the mean direction and $\kappa_m^i \in \mathbb{R}^{+}$ is the concentration parameter.

The parameters $\mu_m^i$ and $\kappa_m^i$ are computed as:
\begin{equation}
\mu_m^i = \frac{f_m(H_m^i)}{\|f_m(H_m^i)\|_2} \in \mathbb{S}^{d-1}, \qquad
\kappa_m^i = \mathrm{softplus}(g_m(H_m^i)) + \epsilon \in \mathbb{R}^{+},
\label{eq:vmf_params_appendix}
\end{equation}
where $f_m(\cdot)$ and $g_m(\cdot)$ are two separate subnetworks used to estimate the directional and concentration parameters, respectively. The normalization ensures that $\mu_m^i$ lies on the unit hypersphere, while the softplus function guarantees that $\kappa_m^i$ remains positive. The constant $\epsilon$ is a small value introduced to ensure numerical stability.

Under this formulation, $\mu_m^i$ represents the principal direction of the latent representation on the hypersphere, while $\kappa_m^i$ controls the dispersion of the distribution around this direction. Intuitively, a larger $\kappa_m^i$ results in a more concentrated distribution, indicating higher certainty, whereas a smaller $\kappa_m^i$ corresponds to a more dispersed distribution, reflecting greater uncertainty. Following prior work~\cite{CIDER2023,ProCo2024,PML2025}, we interpret the concentration parameter $\kappa_m^i$ as a proxy for sample-level reliability of the corresponding modality. This interpretation allows us to quantify the confidence of each modality and later leverage it for reliability-aware fusion.

\paragraph{Probability Density Function.}
The vMF distribution is defined on the unit hypersphere $\mathbb{S}^{d-1} \subset \mathbb{R}^{d}$. The probability density function of a random variable $z \in \mathbb{S}^{d-1}$ is given by:
\begin{equation}
p(z \mid \mu_m^i, \kappa_m^i) = C_d(\kappa_m^i) \exp\left(\kappa_m^i {\mu_m^i}^{\top} z\right),
\end{equation}
where $C_d(\kappa_m^i)$ is the normalization constant defined as:
\begin{equation}
C_d(\kappa_m^i) = \frac{(\kappa_m^i)^{\frac{d}{2}-1}}{(2\pi)^{\frac{d}{2}} I_{\frac{d}{2}-1}(\kappa_m^i)}.
\end{equation}

Here, $I_{\nu}(\cdot)$ denotes the modified Bessel function of the first kind of order $\nu$, which is defined as:
\begin{equation}
I_{\nu}(t) = \sum_{j=0}^{\infty} \frac{\left(\frac{t}{2}\right)^{2j+\nu}}{j! \, \Gamma(j+\nu+1)},
\end{equation}
where $\Gamma(\cdot)$ is the Gamma function.

\paragraph{Discussion.}
The vMF distribution provides a natural way to model directional data on the hypersphere. In our setting, it enables a unified representation that jointly captures both the semantic direction (via $\mu_m^i$) and the uncertainty (via $\kappa_m^i$). This property is important for multimodal learning, where different modalities may exhibit varying levels of reliability across samples. By leveraging the concentration parameter as an uncertainty measure, our framework achieves adaptive, reliability-aware representation learning.


\subsection{Reservoir Sampling Strategy}
\label{appendix: reservoir sampling strategy}

We use reservoir sampling~\cite{vitter1985random} to maintain a fixed-size modality-specific memory bank while avoiding bias toward recently observed samples. For each modality $m$, the memory bank is denoted as
\begin{equation}
\mathcal{B}_m=\{(z_m^r,\mu_m^r,\kappa_m^r,y^r)\}_{r=1}^{N_m},
\end{equation}
where $N_m \leq N_b$ and $N_b$ is the memory bank capacity. During memory construction, samples are sequentially processed by a forward pass without gradient updates. Let $t$ denote the number of samples that have been observed so far for updating the memory bank of modality $m$. For the $t$-th observed sample, we obtain its latent representation $z_m^t$, vMF mean direction $\mu_m^t$, concentration parameter $\kappa_m^t$, and label $y^t$.

If the memory bank is not full, i.e., $t \leq N_b$, the sample is directly inserted into $\mathcal{B}_m$. Once the memory bank reaches its capacity, i.e., $t>N_b$, the new sample is retained with probability $N_b/t$. Specifically, we draw a random integer
\begin{equation}
q \sim \mathrm{Uniform}\{1,2,\dots,t\}.
\end{equation}
If $q \leq N_b$, we replace the $q$-th element in $\mathcal{B}_m$ with the current sample; otherwise, the current sample is discarded. The update rule can be written as:
\begin{equation}
\mathcal{B}_m[q] \leftarrow (z_m^t,\mu_m^t,\kappa_m^t,y^t), 
\quad \text{if } q \leq N_b.
\end{equation}
Otherwise, $\mathcal{B}_m$ remains unchanged.

This strategy guarantees that after observing $t$ samples, each sample has equal probability of being retained in the memory bank:
\begin{equation}
\Pr\left((z_m^s,\mu_m^s,\kappa_m^s,y^s)\in \mathcal{B}_m\right)
=
\min\left(1,\frac{N_b}{t}\right),
\quad \forall s \in \{1,\dots,t\}.
\end{equation}
When $t \leq N_b$, all observed samples are stored. When $t>N_b$, each observed sample is retained with probability $N_b/t$, regardless of its arrival order.

For completeness, we briefly justify this property. For a sample observed at step $s \leq N_b$, it is initially inserted into the memory bank. At any later step $\tau > N_b$, it is replaced only if the new sample is accepted and its stored position is selected. This happens with probability
\begin{equation}
\frac{N_b}{\tau} \cdot \frac{1}{N_b} = \frac{1}{\tau}.
\end{equation}
Therefore, the probability that this sample survives from step $N_b+1$ to step $t$ is
\begin{equation}
\prod_{\tau=N_b+1}^{t}\left(1-\frac{1}{\tau}\right)
=
\prod_{\tau=N_b+1}^{t}\frac{\tau-1}{\tau}
=
\frac{N_b}{t}.
\end{equation}
For a sample observed at step $s>N_b$, it is inserted with probability $N_b/s$. If inserted, it survives each subsequent step $\tau=s+1,\dots,t$ with probability $1-1/\tau$. Hence its final retention probability is
\begin{equation}
\frac{N_b}{s}
\prod_{\tau=s+1}^{t}\left(1-\frac{1}{\tau}\right)
=
\frac{N_b}{s}
\prod_{\tau=s+1}^{t}\frac{\tau-1}{\tau}
=
\frac{N_b}{s}\cdot\frac{s}{t}
=
\frac{N_b}{t}.
\end{equation}
Thus, reservoir sampling maintains a uniformly sampled subset of historical samples under the fixed capacity constraint.

In our implementation, the above procedure is applied independently for each modality-specific memory bank. At the beginning of each training epoch, we rebuild the memory bank by forwarding training samples through the AVIB module without gradient updates and updating $\mathcal{B}_m$ using the reservoir sampling rule. This yields a balanced and order-agnostic memory bank for neighbor retrieval. When the memory bank is fully populated, RME retrieves candidate neighbors from $\mathcal{B}_m$; otherwise, it uses intra-batch candidates as described in Section~\ref{sections_methodology: Reliability-aware Modality Enhancement}.

\section{Experimental Details}

\subsection{Datasets Information}
\label{appendix:datasets_information}
We evaluate the proposed RCE on four benchmark datasets covering three tasks: multimodal sentiment analysis (MSA), multimodal humor detection (MHD), and multimodal sarcasm detection (MSD). We next provide a brief description of each dataset.

\textbf{CMU-MOSI}~\cite{CMU-MOSI2016}: A widely adopted benchmark for MSA, consisting of over 2{,}000 video segments collected from online sources. Each segment is assigned a sentiment intensity score on a seven-point Likert scale ranging from $-3$ (most negative) to $3$ (most positive).
    
\textbf{CMU-MOSEI}~\cite{CMU-MOSEI2018}: One of the largest and most diverse MSA datasets, comprising more than 22{,}000 video segments from over 1{,}000 YouTube speakers across approximately 250 topics. Each segment is annotated with both categorical emotions (six classes) and sentiment scores on the same $-3$ to $3$ scale as CMU-MOSI. In our experiments, we focus on the sentiment annotations for consistency.
    
\textbf{UR-FUNNY}~\cite{UR-FUNNY2019}: A benchmark dataset for MHD, derived from TED talk videos featuring 1,741 speakers. Each target segment, referred to as a punchline, is annotated across language, acoustic, and visual modalities, while preceding segments serve as contextual input. Punchlines are identified using the \textit{laughter} tag in transcripts, which indicates audience laughter; negative samples are constructed when no such cue is present. The dataset is split into 7,614 training, 980 validation, and 994 testing samples. Following prior works~\cite{HKT2021,AtCAF2025,SuCI2025}, we adopt version 2 of UR-FUNNY.
    
\textbf{MUStARD}~\cite{MUStARD2019}: A dataset for MSD, collected from popular television series such as \textit{Friends}, \textit{The Big Bang Theory}, \textit{The Golden Girls}, and \textit{Sarcasmaholics}. It contains 690 video segments labeled as sarcastic or non-sarcastic. Each instance includes the target punchline along with preceding dialogue to provide context.


\subsection{Baselines}
\label{appendix:baselines}
This section offers a thorough overview of the baseline methods utilized in our study.

\textbf{MODS}~\cite{MODS2026} \textcolor{gray}{\footnotesize [AAAI'26]}: Modality Optimization and Dynamic Primary Modality Selection (MODS) tackles modality imbalance by adaptively identifying the most informative modality for each sample. It incorporates a graph-based dynamic sequence compression module to minimize redundancy in non-linguistic modalities, and leverages a primary-modality-focused cross-attention mechanism to strengthen cross-modal integration.

\textbf{PSA-MF}~\cite{PSA-MF2026} \textcolor{gray}{\footnotesize [AAAI'26]}: Personality-Sentiment Aligned Multi-level Fusion (PSA-MF) integrates personality traits into the feature extraction process to learn personalized sentiment representations, and employs a hierarchical fusion strategy to progressively combine multimodal information, thereby enhancing sentiment recognition performance;

\textbf{DecAlign}~\cite{DecAlign2026} \textcolor{gray}{\footnotesize [ICLR'26]}: Decoupled Cross-modal Alignment (DecAlign) tackles multimodal heterogeneity by decomposing representations into modality-specific and shared components. It employs a prototype-guided optimal transport strategy to align modality-specific features and leverages MMD-based distribution matching to enforce cross-modal consistency, further enhanced by a multimodal transformer for high-level fusion.

\textbf{MIB}~\cite{MIB2023} \textcolor{gray}{\footnotesize [TMM'23]}: Multimodal Information Bottleneck (MIB) applies the information bottleneck principle to learn compact, task-relevant representations by reducing redundancy and noise in both unimodal and multimodal features. It provides three variants—E-MIB, L-MIB, and C-MIB—that impose information constraints at different stages of multimodal fusion.

\textbf{ITHP}~\cite{ITHP2024} \textcolor{gray}{\footnotesize [ICLR'24]}: Information-Theoretic Hierarchical Perception (ITHP) leverages the information bottleneck principle by assigning a primary modality and utilizing auxiliary modalities as detectors to extract salient information.

 \textbf{KAN-MCP}~\cite{KAN-MCP2025} \textcolor{gray}{\footnotesize [ACM MM'25]}: Kolmogorov–Arnold Network with Multimodal Clean Pareto (KAN-MCP) combines interpretable cross-modal modeling with DRD-MIB-based feature compression and denoising, enabling discriminative representation learning while mitigating modality imbalance.

\textbf{OMIB}~\cite{OMIB2025} \textcolor{gray}{\footnotesize [ICML'25]}: Optimal Multimodal Information Bottleneck (OMIB) extends the MIB framework by theoretically constraining regularization weights to ensure an optimal bottleneck representation, while dynamically adjusting modality-specific regularization to alleviate modality imbalance.

\textbf{QMF}~\cite{QMF2023} \textcolor{gray}{\footnotesize [ICML'23]}: Quality-aware Multimodal Fusion (QMF) provides a theoretically grounded framework for robust multimodal learning by analyzing multimodal fusion from a generalization perspective. It leverages uncertainty estimation to dynamically assess modality quality, enabling adaptive cross-modal interaction and mitigating the impact of low-quality data, thereby promoting more reliable and quality-aware feature integration.

\textbf{PML}~\cite{PML2025} \textcolor{gray}{\footnotesize [IJCAI'25]}: Probabilistic Multimodal Learning (PML) models representations with von Mises–Fisher (vMF) distributions to capture intrinsic uncertainty and learn reliable directional features. It employs a vMF-based prototypical contrastive learning paradigm to enhance class discrimination, and introduces a reliability-aware fusion mechanism to dynamically regulate modality contributions and resolve cross-modal conflicts.

\textbf{HME}~\cite{HME2025} \textcolor{gray}{\footnotesize [NeurIPS'25]}: Hyper-Modality Enhancement (HME) tackles missing modality scenarios by enriching observed modalities with semantically relevant cues from other samples, avoiding explicit reconstruction. It further adopts an uncertainty-aware fusion strategy to adaptively balance original and enhanced representations, improving robustness to incomplete and noisy inputs.

\textbf{CyIN}~\cite{CyIN2025} \textcolor{gray}{\footnotesize [NeurIPS'25]}: Cyclic INformative Learning (CyIN) addresses dynamic missing modality scenarios by constructing an informative latent space via token- and label-level Information Bottleneck (IB) applied cyclically across modalities. It further introduces cross-modal cyclic translation to reconstruct missing modalities through forward and reverse processes, enabling unified optimization for both complete and incomplete multimodal learning while enhancing cross-modal interaction and fusion.

\textbf{Self-MM}~\cite{Self-MM2021} \textcolor{gray}{\footnotesize [AAAI'21]}: Self-Supervised Multi-task Multimodal (Self-MM) sentiment analysis framework generates pseudo labels for each modality based on annotated global sentiment labels, enabling the learning of more discriminative unimodal representations.

\textbf{FDMER}~\cite{FDMER2022} \textcolor{gray}{\footnotesize [ACM MM'22]}: Feature-Disentangled Multimodal Emotion Recognition (FDMER) alleviates distribution gaps and redundancy across modalities by decomposing features into modality-invariant and modality-specific subspaces. It employs common and private encoders with adversarial training to enforce consistency and diversity, and utilizes a cross-modal attention fusion module to learn adaptive multimodal representations.

\textbf{DMD}~\cite{DMD2023} \textcolor{gray}{\footnotesize [CVPR'23]}: Decoupled Multimodal Distillation (DMD) improves emotion recognition by separating each modality into modality-relevant and modality-exclusive components, and conducting adaptive cross-modal knowledge distillation through a dynamic graph structure.

\textbf{AGM}~\cite{AGM2023} \textcolor{gray}{\footnotesize [ICCV'23]}: Adaptive Gradient Modulation (AGM) addresses modality competition in multimodal learning by adaptively adjusting gradients during training to balance modality contributions across various fusion strategies. It further introduces a novel metric based on the mono-modal concept to quantify competition strength, providing insights into modality dominance and improving overall model performance.

\textbf{AtCAF}~\cite{AtCAF2025} \textcolor{gray}{\footnotesize [Information Fusion'25]}: Attention-based Causality-Aware Fusion (AtCAF) learns causality-aware multimodal representations through a text debiasing module and counterfactual cross-modal attention for sentiment analysis.

\textbf{SuCI}~\citep{SuCI2025} \textcolor{gray}{\footnotesize [AAAI'25]}: Subject Causal Intervention (SuCI) mitigates subject-related spurious correlations by modeling subjects as confounders and applying causal intervention, leading to more robust multimodal language understanding and improved cross-subject generalization.

\textbf{DHMD}~\cite{DHMD2026} \textcolor{gray}{\footnotesize [TPAMI'26]}: Decoupled Hierarchical Multimodal Distillation (DHMD) separates multimodal features into modality-invariant and modality-specific components via a self-regression mechanism, and adopts a hierarchical distillation framework—combining graph-based coarse-grained and dictionary-based fine-grained distillation—to improve cross-modal alignment and emotion recognition.

\textbf{MAG}~\cite{MAG2020} \textcolor{gray}{\footnotesize [ACL'20]}: Multimodal Adaptation Gate (MAG) introduces an adaptation gate that enables large pre-trained transformers to incorporate multimodal signals during the fine-tuning stage.

\textbf{BBFN}~\cite{BBFN2021} \textcolor{gray}{\footnotesize [ICMI'21]}: Bi-Bimodal Fusion Network (BBFN) conducts both fusion and separation over pairwise modality representations, and utilizes a gated Transformer to alleviate modality imbalance.

\textbf{MuLOT}~\cite{MuLOT2022} \textcolor{gray}{\footnotesize [WACV'22]}: Multimodal Learning using Optimal Transport (MuLOT) addresses resource-constrained sarcasm and humor detection by leveraging self-attention to model intra-modal relationships and optimal transport to capture cross-modal alignment, followed by multimodal attention fusion to integrate complementary information across modalities.

\textbf{MIL}~\cite{MIL2023} \textcolor{gray}{\footnotesize [TAI'23]}: Multimodal Interaction Learning (MIL) is a multitask framework for joint sarcasm and sentiment detection that models both shared and task-specific features. It introduces a cross-modal target attention mechanism to enhance multimodal fusion and facilitate interaction between related tasks.

\textbf{MOAC}~\cite{MOAC2025} \textcolor{gray}{\footnotesize [WWW'25]}: Multimodal Ordinal Affective Computing (MOAC) improves affective modeling by incorporating ordinal learning at both the label level (coarse-grained) and feature level (fine-grained) within multimodal representations.


\subsection{Feature Extraction}
\label{appendix:feature_extraction}

\textbf{Textual Modality:} For the CMU-MOSI and CMU-MOSEI datasets, we adopt DeBERTa~\cite{DeBERTa2020} to extract high-level textual representations, following prior work. For the UR-FUNNY dataset, ALBERT~\cite{ALBERT2019} is employed as the text encoder. Specifically, the context and punchline token sequences are concatenated to form the input sequence: $U_l = C_l \oplus \texttt{[SEP]} \oplus P_l$, where \texttt{[SEP]} is used to separate the context tokens $C_l$ and punchline tokens $P_l$. For the MUStARD dataset, contextual word representations are extracted using a pretrained BERT~\cite{BERT2019} model. 

For the experiments reported in Figure~\ref{fig:random_missing} of the main paper, in order to ensure a fair comparison with CyIN~\cite{CyIN2025}, both RCE and HME~\cite{HME2025} adopt BERT~\cite{BERT2019} to extract textual modality features.

\textbf{Acoustic Modality: } Acoustic features for CMU-MOSI, CMU-MOSEI, UR-FUNNY, and MUStARD are extracted using COVAREP~\cite{COVAREP2014}, including 12-dimensional MFCCs, pitch, speech polarity, glottal closure instants, and spectral envelope descriptors. These features are computed over the full audio segment of each utterance, forming temporal sequences that capture dynamic vocal variations.

\textbf{Visual Modality: } For CMU-MOSI and CMU-MOSEI, visual features are obtained using Facet (iMotions 2017, \url{https://imotions.com/}), including facial action units, landmarks, head pose, and other expression-related cues, organized as temporal sequences to capture facial dynamics. For UR-FUNNY and MUStARD, OpenFace~\cite{OpenFace2.02016} is used to extract facial action units along with both rigid and non-rigid facial shape parameters.

For the CMU-MOSI dataset, the feature dimensions of the textual, acoustic, and visual modalities are 768, 74, and 47, respectively. For CMU-MOSEI, the corresponding dimensions are 768, 74, and 35. For both UR-FUNNY and MUStARD, the dimensions of the textual, acoustic, and visual modalities are 768, 60, and 36, respectively. In addition, UR-FUNNY includes an extra humor centric feature (HCF) modality with a dimension of 4. Details of the HCF feature extraction process can be found in~\cite{HKT2021}.


\subsection{Implementation Details}
\label{appendix:implementation_details}

We implement RCE using the PyTorch framework and conduct all experiments on a single NVIDIA GeForce RTX 4090 GPU, using CUDA 12.8 and PyTorch 2.8.0. The model is optimized using AdamW~\cite{AdamW} with linear warm-up. Detailed hyper-parameter settings for different datasets are summarized in Table~\ref{tab:hyper_parameters}.

Several hyper-parameters are shared across all datasets. Specifically, warm-up is applied in all experiments, and AdamW is consistently adopted as the optimizer. The modality enhancement process uses a modality-specific memory bank with capacity $N_b$, where neighbor retrieval is performed based on the top-$k$ ratio $\rho$. The prompt length is denoted by $l_p$, the label penalty weight by $\alpha$, the AVIB trade-off coefficient by $\beta$, and the overall loss balancing weight by $\lambda$.

For the MSA datasets CMU-MOSI and CMU-MOSEI, we use relatively larger feature dimensions and memory bank capacities to capture richer multimodal contextual information. Specifically, the feature dimensions are set to $d=256$ and $d=192$, respectively, while the memory bank capacities are set to $1024$ and $2048$. The batch sizes are set to $48$ for CMU-MOSI and $128$ for CMU-MOSEI, with initial learning rates of $2\times10^{-5}$ and $3\times10^{-5}$, respectively. Both datasets use a dropout rate of $0.4$ and AVIB trade-off coefficient $\beta=1\mathrm{e}{-3}$. For the MHD and MSD datasets UR-FUNNY and MUStARD, we adopt comparatively smaller dropout rates to preserve discriminative multimodal cues. The feature dimensions are set to $d=192$ and $d=144$, respectively. The prompt lengths differ across datasets, with $l_p=3$ for UR-FUNNY and $l_p=8$ for MUStARD. Both datasets use a memory bank capacity of $1024$, while the top-$k$ ratios are set to $0.3$ and $0.5$, respectively. The AVIB trade-off coefficients are set to $1\mathrm{e}{-4}$ for UR-FUNNY and $1\mathrm{e}{-2}$ for MUStARD.

To determine the optimal configuration, we perform a grid search with thirty random trials on the validation set. The batch size is selected from ${32,48,64,128,256}$, the initial learning rate is searched over ${1\mathrm{e}{-6},3\mathrm{e}{-6},8\mathrm{e}{-6},1\mathrm{e}{-5},2\mathrm{e}{-5},3\mathrm{e}{-5}}$, and the feature dimension $d$ is selected from ${96,144,192,256}$. The dropout rate is tuned within ${0.1,0.2,0.3,0.4,0.5}$, the prompt length $l_p$ is searched over ${2,3,4,6,8}$, and the memory bank capacity $N_b$ is selected from ${256,512,1024,2048}$. In addition, the top-$k$ ratio $\rho$ is tuned within ${0.1,0.2,\dots,0.7}$, while the label penalty weight $\alpha$ and loss balancing weight $\lambda$ are tuned within ${0.1,0.2,\dots,1.0}$. The AVIB trade-off coefficient $\beta$ is searched over ${1\mathrm{e}{-2},1\mathrm{e}{-3},1\mathrm{e}{-4},1\mathrm{e}{-5}}$. The final model is selected based on the configuration that achieves the lowest \textit{MAE} on the validation set.

Notably, for a fair comparison, we re-implement C-MIB~\cite{MIB2023}, ITHP~\cite{ITHP2024}, KAN-MCP~\cite{KAN-MCP2025}, OMIB~\cite{OMIB2025}, QMF~\cite{QMF2023}, PML~\cite{PML2025}, and HME~\cite{HME2025} using the same feature extraction strategy as RCE. The implementations are based on their publicly available official codebases, and the hyper-parameter search is conducted according to the parameter ranges reported in their original papers using thirty random search trials on the validation set to determine the optimal configuration. For the remaining baselines, unless otherwise specified, the reported results are directly taken from the corresponding original papers.

For the MHD task, the results of Self-MM~\cite{Self-MM2021}, FDMER~\cite{FDMER2022}, and DMD~\cite{DMD2023} are taken from the SuCI~\cite{SuCI2025} paper. For the MSD task, the results of MAG-XLNet~\cite{MAG2020} and BBFN~\cite{BBFN2021} are taken from the MuLOT~\cite{MuLOT2022} paper. Results for all other baselines are directly taken from their corresponding original papers.

\begin{table*}[t]
\caption{Hyper-parameter settings of RCE across different datasets.}
\centering
\resizebox{0.9\linewidth}{!}{
\begin{tabular}{c c c c c}
\toprule[1.5pt]

\textbf{Hyper-parameter} & \textbf{CMU-MOSI} & \textbf{CMU-MOSEI} & \textbf{UR-FUNNY} & \textbf{MUStARD} \\

\midrule

Batch Size & 48 & 128 & 256 & 32 \\
Epochs & 50 & 10 & 10 & 20 \\
Warm-up & \checkmark & \checkmark & \checkmark & \checkmark \\
Initial Learning Rate & $2 \times 10^{-5}$ & $3 \times 10^{-5}$ & $1 \times 10^{-5}$ & $3 \times 10^{-6}$ \\
Optimizer & AdamW & AdamW & AdamW & AdamW \\
Dropout Rate & 0.4 & 0.4 & 0.1 & 0.1 \\
Feature Dimension $d$ & 256 & 192 & 192 & 144 \\
Prompt Length $l_p$ & 4 & 6 & 3 & 8 \\
Memory Bank Capacity $N_b$ & 1024 & 2048 & 1024 & 1024 \\
Top-$k$ Ratio $\rho$ & 0.6 & 0.4 & 0.3 & 0.5 \\
Label Penalty Weight $\alpha$ & 0.4 & 0.1 & 0.3 & 0.3 \\
AVIB Trade-off Coefficient $\beta$ & 1e-3 & 1e-3 & 1e-4 & 1e-2 \\
Loss Balancing Weight $\lambda$& 0.7 & 1.0 & 0.6 & 0.4 \\

\bottomrule[1.5pt]
\end{tabular}
}
\label{tab:hyper_parameters}
\end{table*}


\subsection{Evaluation Metrics}
\label{appendix:evaluation_metrics}

We evaluate the model’s performance on the MSA task using a set of well-established metrics, reported for both CMU-MOSI and CMU-MOSEI datasets. For interpretability, classification results are presented as percentages. These metrics are calculated as follows:

\textbf{Seven-category Classification Accuracy (\textit{Acc7}):} Measures the model’s ability to predict fine-grained sentiment categories by dividing the sentiment score range ($-3$ to $3$) into seven equal intervals. The metric is defined as:
\begin{equation}
\mathrm{Acc7} = \frac{1}{n}\sum_{i=1}^{n}\mathbf{1}\big(\hat{c}_i = c_i\big),
\end{equation}
where $c_i$ and $\hat{c}_i$ denote the ground-truth and predicted categories of sample $i$, respectively, and $\mathbf{1}(\cdot)$ is the indicator function. Higher values indicate better fine-grained sentiment classification.

\textbf{Binary Classification Accuracy (\textit{Acc2}):} Following prior work~\cite{ITHP2024,KAN-MCP2025}, we report binary sentiment classification results by distinguishing negative ($<0$) and positive ($>0$) samples, excluding neutral cases. The metric is formulated as:  
\begin{equation}
\mathrm{Acc2} = \frac{TP + TN}{TP + TN + FP + FN},
\end{equation}
where $TP$, $TN$, $FP$, and $FN$ denote true positives, true negatives, false positives, and false negatives, respectively.

\textbf{Weighted F1-score (\textit{F1}):} Computes the harmonic mean of precision and recall while considering class-specific weights to mitigate imbalance. It is formulated as:  
\begin{equation}
\mathrm{F1} = 2 \cdot \frac{Precision \cdot Recall}{Precision + Recall}, 
\end{equation}
where $\mathrm{Precision} = \tfrac{TP}{TP+FP}$ and $\mathrm{Recall} = \tfrac{TP}{TP+FN}$.

\textbf{Mean Absolute Error (\textit{MAE}):} Represents the average magnitude of prediction errors with respect to the ground-truth sentiment scores. It directly corresponds to the original sentiment scale, making it both intuitive and informative:  
\begin{equation}
\mathrm{MAE}(\hat{y}, y) = \frac{1}{n}\sum_{i=1}^{n}|\hat{y}_i - y_i|,
\end{equation}
where $y_i$ is the true label, $\hat{y}_i$ is the predicted value, and $n$ is the total number of predictions.

\textbf{Pearson Correlation Coefficient (\textit{Corr}):} Quantifies the strength and direction of the linear relationship between predicted and true sentiment scores:  
\begin{equation}
\mathrm{Corr}(x, y) = \frac{\sum_{i=1}^{n}(x_i - \bar{x})(y_i - \bar{y})}{\sqrt{\sum_{i=1}^{n}(x_i - \bar{x})^2}\sqrt{\sum_{i=1}^{n}(y_i - \bar{y})^2}},
\end{equation}
where $x_i$ and $y_i$ denote predicted and ground-truth values, respectively, and $\bar{x}, \bar{y}$ are their means.

For the MHD and MSD tasks, following prior work~\cite{SuCI2025,AtCAF2025,MOAC2025,DHMD2026}, we report binary accuracy only, which measures the model’s capability to differentiate between humorous and non-humorous, as well as sarcastic and non-sarcastic, instances.

\section{More Experimental Results}
\label{appendix:additional_experimental_results}

\begin{table}[h]
\centering  
\setlength{\belowcaptionskip}{10pt}
\renewcommand\arraystretch{1.1}
\caption{Performance comparison of RCE and baseline methods under fixed missing protocol.}

\label{tab:test_missing}
\resizebox{\textwidth}{!}{

\begin{tabular}{lccccccccccc}

\toprule[1.5pt]

\multirow{2}{*}{\textbf{Method}} 
& \multicolumn{5}{c}{\textbf{CMU-MOSI}} & & \multicolumn{5}{c}{\textbf{CMU-MOSEI}} \\
\cmidrule{2-6} \cmidrule{8-12}
& \textit{Acc7\%$\uparrow$} & \textit{Acc2\%$\uparrow$} & \textit{F1\%$\uparrow$} & \textit{MAE$\downarrow$} &\textit{Corr$\uparrow$} & & \textit{Acc7\%$\uparrow$} & \textit{Acc2\%$\uparrow$} & \textit{F1\%$\uparrow$} & \textit{MAE$\downarrow$} &\textit{Corr$\uparrow$} \\

\midrule 
\rowcolor{offlinecolor} \multicolumn{12}{c}{\textit{\textbf{Testing Condition: \textnormal{\{}t\textnormal{\}}}}} \\ 
\midrule

C-MIB~\cite{MIB2023} 
& 45.69  & 87.33 & 87.28 & 0.644 & 0.842 &
& 51.76  & 85.66 & 85.62 & 0.550 & 0.771 \\

ITHP~\cite{ITHP2024} 
& 43.50  & 86.72 & 86.71 & 0.678 & 0.838 &
& 52.40  & 86.96 & 86.94 & 0.541 & 0.791 \\

KAN-MCP~\cite{KAN-MCP2025} 
& 45.69  & 88.24 & 88.21 & 0.635 & 0.851 &
& 53.87  & 86.80 & 86.80 & 0.531 & 0.774 \\

OMIB~\cite{OMIB2025} 
& \best{49.49}  & 87.63 & 87.51 & 0.632 & 0.844 &
& 53.37  & \best{87.18} & 87.14 & 0.529 & 0.786 \\

QMF~\cite{QMF2023} 
& 42.92  & 87.63 & 87.65 & 0.691 & 0.846 &
& 51.32  & 85.75 & 85.31 & 0.547 & 0.789 \\

PML~\cite{PML2025} 
& 45.99  & 87.94 & 87.86 & 0.657 & 0.837 &
& 53.84  & 86.99 & 86.94 & 0.526 & 0.787 \\

HME~\cite{HME2025} 
& 41.17  & 88.55 & 88.49 & 0.710 & 0.856 &
& 54.08  & 87.15 & \best{87.15} & 0.518 & 0.787 \\

\textbf{RCE (Ours)} 
& 46.72  & \best{89.01} & \best{88.94} & \best{0.615} & \best{0.860} &
& \best{55.18}  & 86.96 & 86.93 & \best{0.505} & \best{0.795} \\

\midrule 
\rowcolor{offlinecolor} \multicolumn{12}{c}{\textit{\textbf{Testing Condition: \textnormal{\{}t, a\textnormal{\}}}}} \\ 
\midrule

C-MIB~\cite{MIB2023} 
& 43.94  & 84.73 & 84.74 & 0.725 & 0.804 &
& 51.24  & 85.50 & 85.48 & 0.560 & 0.760 \\

ITHP~\cite{ITHP2024} 
& 44.53  & 86.87 & 86.87 & 0.689 & 0.838 &
& 53.54  & 86.60 & 86.71 & 0.541 & 0.787 \\

KAN-MCP~\cite{KAN-MCP2025} 
& 47.15  & 86.87 & 86.84 & 0.623 & 0.853 &
& 53.46  & 86.85 & 86.90 & 0.526 & 0.788 \\

OMIB~\cite{OMIB2025} 
& 48.32  & 88.40 & 88.32 & 0.633 & 0.844 &
& 53.37  & 87.15 & 87.11 & 0.529 & 0.786 \\

QMF~\cite{QMF2023} 
& 45.69  & 88.09 & 88.04 & 0.653 & 0.846 &
& 52.10  & 86.93 & 86.72 & 0.532 & 0.790 \\

PML~\cite{PML2025} 
& 46.28  & 87.94 & 87.86 & 0.656 & 0.837 &
& 53.50  & 86.99 & 86.92 & 0.530 & 0.787 \\

HME~\cite{HME2025} 
& 45.99  & 88.70 & 88.70 & 0.620 & 0.856 &
& 54.08  & 87.13 & 86.95 & 0.519 & 0.793 \\

\textbf{RCE (Ours)} 
& \best{48.61}  & \best{89.16} & \best{89.11} & \best{0.596} & \best{0.860} &
& \best{55.27}  & \best{87.40} & \best{87.44} & \best{0.502} & \best{0.800} \\

\midrule 
\rowcolor{offlinecolor} \multicolumn{12}{c}{\textit{\textbf{Testing Condition: \textnormal{\{}t, v\textnormal{\}}}}} \\ 
\midrule

C-MIB~\cite{MIB2023} 
& 47.30  & 87.63 & 87.62 & 0.651 & 0.840 &
& 51.30  & 86.33 & 86.31 & 0.561 & 0.768 \\

ITHP~\cite{ITHP2024} 
& 44.09  & 84.89 & 84.90 & 0.722 & 0.816 &
& 52.25  & 86.35 & 86.46 & 0.550 & 0.792 \\

KAN-MCP~\cite{KAN-MCP2025} 
& 45.55  & 87.63 & 87.60 & 0.668 & 0.834 &
& 53.03  & 86.57 & 86.66 & 0.528 & 0.784 \\

OMIB~\cite{OMIB2025} 
& \best{48.61}  & 88.09 & 88.00 & 0.633 & 0.844 &
& 53.52  & 87.24 & 87.21 & 0.527 & 0.786 \\

QMF~\cite{QMF2023} 
& 45.11  & 87.79 & 87.80 & 0.665 & 0.847 &
& 52.94  & 86.99 & 86.77 & 0.529 & 0.790 \\

PML~\cite{PML2025} 
& 46.42  & 87.94 & 87.86 & 0.656 & 0.837 &
& 53.65  & 87.15 & 87.15 & 0.528 & 0.786 \\

HME~\cite{HME2025} 
& 44.09  & \best{89.16} & \best{89.14} & 0.653 & 0.856 &
& 54.06  & 85.47 & 85.64 & 0.519 & 0.792 \\

\textbf{RCE (Ours)} 
& 48.47  & 89.01 & 88.93 & \best{0.592} & \best{0.861} &
& \best{54.92}  & \best{87.49} & \best{87.39} & \best{0.503} & \best{0.800} \\

\bottomrule[1.5pt]

\end{tabular}
}
\end{table}

\subsection{Results under Fixed Missing Protocols}
\label{appendix:results_fixed_missing}

Under the fixed missing protocol, we further evaluate RCE against a wide range of competitive missing-modality methods, including C-MIB~\cite{MIB2023}, ITHP~\cite{ITHP2024}, KAN-MCP~\cite{KAN-MCP2025}, OMIB~\cite{OMIB2025}, QMF~\cite{QMF2023}, PML~\cite{PML2025}, and HME~\cite{HME2025}. As reported in Table~\ref{tab:test_missing}, RCE achieves the best overall performance across most fixed modality-availability settings. Specifically, when only the textual modality is available, RCE obtains the best results on four out of five metrics on CMU-MOSI and three out of five metrics on CMU-MOSEI, showing that the proposed method can still produce reliable predictions under highly sparse modality conditions. When both text and audio are available, RCE consistently outperforms all baselines on both datasets across all evaluation metrics, demonstrating its strong ability to exploit complementary acoustic cues in addition to the dominant textual modality.

For the setting where text and vision are available, RCE also exhibits highly competitive and stable performance. On CMU-MOSEI, it achieves the best results on all five metrics, while on CMU-MOSI it obtains the lowest $MAE$ and highest correlation, with $Acc2$ and $F1$ remaining very close to the best-performing baseline. These results suggest that RCE is not limited to a specific modality combination, but can adaptively benefit from different observed modalities. Overall, under the fixed missing protocol, RCE achieves the best performance on 24 out of 30 metric comparisons, validating its robustness to diverse deterministic missing patterns. The improvements can be attributed to the reliability-aware modality enhancement mechanism, which retrieves informative cross-sample cues from the memory bank, and the multilevel fusion strategy, which adaptively integrates unimodal, enhanced, and cross-modal representations according to modality reliability.


\subsection{Results under Noisy Modalities}
\label{appendix:results_noise_modalities}

In addition to missing-modality evaluation, we conduct noise robustness experiments to further examine whether RCE can handle corrupted but still observable modalities.

\begin{table}[h]
\centering  
\setlength{\belowcaptionskip}{10pt}
\renewcommand\arraystretch{1.1}
\caption{Comparison under different Gaussian noise intensities on CMU-MOSI and CMU-MOSEI.}

\label{tab:noise_gaussian}
\resizebox{\textwidth}{!}{

\begin{tabular}{lccccccccccc}

\toprule[1.5pt]

\multirow{2}{*}{\textbf{Method}} 
& \multicolumn{5}{c}{\textbf{CMU-MOSI}} & & \multicolumn{5}{c}{\textbf{CMU-MOSEI}} \\
\cmidrule{2-6} \cmidrule{8-12}
& \textit{Acc7\%$\uparrow$} & \textit{Acc2\%$\uparrow$} & \textit{F1\%$\uparrow$} & \textit{MAE$\downarrow$} &\textit{Corr$\uparrow$} & & \textit{Acc7\%$\uparrow$} & \textit{Acc2\%$\uparrow$} & \textit{F1\%$\uparrow$} & \textit{MAE$\downarrow$} &\textit{Corr$\uparrow$} \\

\midrule 
\rowcolor{offlinecolor} \multicolumn{12}{c}{\textit{\textbf{Gaussian Noise Intensity: 1.0}}} \\ 
\midrule

C-MIB~\cite{MIB2023}
& 42.34  & 85.65 & 85.68 & 0.750 & 0.782 &
& 51.07  & 84.50 & 84.65 & 0.560 & 0.766 \\

ITHP~\cite{ITHP2024} 
& 45.55  & 86.11 & 86.10 & 0.697 & \best{0.822} &
& 52.10  & 84.03 & 84.27 & 0.564 & 0.781 \\

KAN-MCP~\cite{KAN-MCP2025} 
& 45.99  & 85.80 & 85.76 & 0.710 & 0.805 &
& 52.04  & \best{86.88} & \best{86.84} & 0.543 & 0.775 \\

OMIB~\cite{OMIB2025} 
& 42.34  & 83.51 & 83.51 & 0.751 & 0.785 &
& 50.36  & 84.42 & 84.63 & 0.562 & 0.770 \\

QMF~\cite{QMF2023} 
& 44.96  & 86.41 & 86.36 & 0.691 & 0.809 &
& 50.81  & 85.86 & 85.78 & 0.554 & 0.762 \\

PML~\cite{PML2025} 
& 44.38  & 86.41 & 86.42 & 0.732 & 0.795 &
& 52.62  & 85.94 & 85.85 & 0.547 & 0.766 \\

HME~\cite{HME2025} 
& 45.69  & 86.56 & 86.55 & 0.677 & 0.813 &
& 54.08  & 85.88 & 85.94 & 0.527 & 0.778 \\

\textbf{RCE (Ours)} 
& \best{46.42}  & \best{87.02} & \best{86.88} & \best{0.676} & 0.815 &
& \best{54.12}  & 86.80 & 86.76 & \best{0.518} & \best{0.783} \\

\midrule 
\rowcolor{offlinecolor} \multicolumn{12}{c}{\textit{\textbf{Gaussian Noise Intensity: 5.0}}} \\ 
\midrule

C-MIB~\cite{MIB2023} 
& 28.47  & 76.18 & 76.27 & 1.063 & 0.578 &
& 48.29  & 82.96 & 82.75 & 0.616 & 0.695 \\

ITHP~\cite{ITHP2024} 
& 37.08  & 79.24 & 79.34 & 0.908 & 0.683 &
& 49.06  & 82.60 & 82.55 & 0.602 & \best{0.717} \\

KAN-MCP~\cite{KAN-MCP2025} 
& 39.56  & 77.71 & 77.84 & 0.923 & 0.677 &
& 50.23  & 82.54 & 82.44 & 0.596 & 0.685 \\

OMIB~\cite{OMIB2025} 
& 39.42  & 79.08 & 79.20 & 0.859 & 0.716 &
& 50.08  & 82.68 & 82.20 & 0.588 & 0.706 \\

QMF~\cite{QMF2023} 
& 33.43  & 77.56 & 77.58 & 0.948 & 0.678 &
& 46.56  & 83.45 & 82.97 & 0.615 & 0.716 \\

PML~\cite{PML2025} 
& 36.93  & 79.24 & 79.32 & 0.880 & 0.713 &
& 51.30  & 83.20 & 83.18 & 0.581 & 0.715 \\

HME~\cite{HME2025} 
& \best{42.77}  & 80.31 & 80.37 & \best{0.801} & \best{0.719} &
& \best{51.54}  & 80.19 & 80.50 & 0.585 & 0.721 \\

\textbf{RCE (Ours)} 
& 38.83  & \best{80.76} & \best{80.74} & 0.886 & 0.689 &
& 51.00  & \best{83.51} & \best{83.36} & \best{0.577} & 0.715 \\

\midrule 
\rowcolor{offlinecolor} \multicolumn{12}{c}{\textit{\textbf{Gaussian Noise Intensity: 10.0}}} \\ 
\midrule

C-MIB~\cite{MIB2023} 
& 15.62  & 41.98 & 25.28 & 1.466 & -0.010 &
& 45.55  & 73.54 & 72.01 & 0.707 & 0.546 \\

ITHP~\cite{ITHP2024} 
& 28.91  & 69.31 & 69.14 & 1.081 & 0.562 &
& 46.00  & 74.89 & 73.27 & 0.702 & 0.567 \\

KAN-MCP~\cite{KAN-MCP2025} 
& 14.74  & 42.29 & 25.14 & 1.493 & -0.004 &
& 47.02  & 75.11 & \best{73.85} & 0.683 & \best{0.567} \\

OMIB~\cite{OMIB2025} 
& 15.47  & 43.05 & 27.04 & 1.462 & 0.062 &
& 45.75  & 73.62 & 71.95 & 0.705 & 0.535 \\

QMF~\cite{QMF2023} 
& 26.28  & 68.09 & 68.23 & 1.107 & 0.538 &
& 46.87  & 74.34 & 71.80 & 0.688 & 0.561 \\

PML~\cite{PML2025} 
& 15.33  & 44.12 & 29.96 & 1.441 & 0.158 &
& 47.36  & 75.39 & 73.77 & 0.687 & 0.555 \\

HME~\cite{HME2025} 
& 32.26  & 67.48 & 66.45 & 1.039 & 0.581 &
& 47.79  & 74.72 & 73.26 & 0.678 & 0.559 \\

\textbf{RCE (Ours)} 
& \best{32.85}  & \best{72.82} & \best{72.61} & \best{0.983} & \best{0.625} &
& \best{48.09}  & \best{75.41} & 72.50 & \best{0.678} & 0.560 \\

\bottomrule[1.5pt]

\end{tabular}
}
\end{table}

\paragraph{Noise setting.}
Following the noise evaluation protocols used in EAU~\cite{EAU2024} and QMF~\cite{QMF2023}, we further examine the robustness of RCE under corrupted multimodal inputs. Since the original noise operations are mainly designed for image-like data, we adapt them to multimodal sentiment analysis with text tokens and continuous acoustic/visual feature sequences. For the textual modality, noise is directly applied to the encoded \texttt{input\_ids} by replacing selected token ids, which minimally changes the preprocessing pipeline. Textual noise can be applied to the train, validation, and test splits. For acoustic and visual modalities, we adapt Gaussian and Salt-pepper noise to continuous feature sequences. Specifically, Gaussian noise is not generated using image-style amplitude scaling; instead, the perturbation scale is determined by the standard deviation of valid features in each sample multiplied by $\texttt{noise}/10$. For Salt-pepper noise, rather than replacing pixels with 0 or 255, we replace selected valid feature elements with the minimum or maximum value of the corresponding sample. Acoustic and visual noise is applied only to the test split to evaluate robustness under corrupted inference-time observations. The noise intensity parameter \texttt{noise} ranges from 0 to 10, where larger values indicate stronger corruption. For text, after a sample-level trigger with probability 0.5, each token is replaced with probability $\texttt{noise}/10$. For acoustic and visual Gaussian noise, a sample-level trigger with probability 0.5 determines whether noise is added, and the perturbation magnitude is proportional to $\texttt{noise}/10$. For acoustic and visual Salt-pepper noise, the corruption process is first triggered with probability 0.5, and the corruption strength is further controlled by $\texttt{noise}/10$.

\begin{table}[t]
\centering  
\setlength{\belowcaptionskip}{10pt}
\renewcommand\arraystretch{1.1}
\caption{Comparison under different Salt-pepper noise intensities on CMU-MOSI and CMU-MOSEI.}

\label{tab:noise_salt}
\resizebox{\textwidth}{!}{

\begin{tabular}{lccccccccccc}

\toprule[1.5pt]

\multirow{2}{*}{\textbf{Method}} 
& \multicolumn{5}{c}{\textbf{CMU-MOSI}} & & \multicolumn{5}{c}{\textbf{CMU-MOSEI}} \\
\cmidrule{2-6} \cmidrule{8-12}
& \textit{Acc7\%$\uparrow$} & \textit{Acc2\%$\uparrow$} & \textit{F1\%$\uparrow$} & \textit{MAE$\downarrow$} &\textit{Corr$\uparrow$} & & \textit{Acc7\%$\uparrow$} & \textit{Acc2\%$\uparrow$} & \textit{F1\%$\uparrow$} & \textit{MAE$\downarrow$} &\textit{Corr$\uparrow$} \\

\midrule 
\rowcolor{offlinecolor} \multicolumn{12}{c}{\textit{\textbf{Salt-pepper Noise Intensity: 1.0}}} \\ 
\midrule

C-MIB 
& 42.04  & 85.80 & 85.80 & 0.751 & 0.786 &
& 51.93  & 85.52 & 85.41 & 0.555 & 0.765 \\

ITHP 
& 15.47  & 42.29 & 25.14 & 1.471 & -0.030 &
& 52.49  & 85.75 & 85.71 & 0.552 & 0.777 \\

KAN-MCP 
& 48.47  & 87.02 & 87.02 &  0.657 &  \best{0.835} &
& 52.98  & 86.85 & 86.90 & 0.531 & 0.777 \\

OMIB 
& 37.52  & 78.32 & 78.43 & 0.903 & 0.691 &
& 53.24  & 86.16 & 86.24 & 0.533 & 0.776 \\

QMF 
& 43.65  & 84.58 & 84.60 & 0.735 & 0.789 &
& 48.24  & 83.65 & 83.56 & 0.602 & 0.706 \\

PML 
& 43.50  & 85.95 & 85.89 & 0.721 & 0.802 &
& 52.23  & 85.14 & 85.03 & 0.545 & 0.763 \\

HME 
& 45.99  & 86.87 & 86.88 & 0.657 & 0.830 &
& 53.46  & 86.63 & 86.55 & \best{0.524} & 0.776 \\

\textbf{RCE (Ours)} 
& \best{48.47}  &  \best{87.79} &  \best{87.84} & \best{0.634} & 0.829 &
& \best{54.04}  & \best{86.99} & \best{86.99} & 0.527 & \best{0.779} \\

\midrule 
\rowcolor{offlinecolor} \multicolumn{12}{c}{\textit{\textbf{Salt-pepper Noise Intensity: 5.0}}} \\ 
\midrule

C-MIB 
& 39.42  & 78.63 & 78.75 & 0.927 & 0.661 &
& 48.07  & 83.40 & 83.34 & 0.610 & 0.715 \\

ITHP 
& 14.31  & 46.26 & 36.20 & 1.472 & 0.118 &
& 49.97  & 83.07 & 83.21 & 0.601 & 0.722 \\

KAN-MCP 
& 31.68  & 76.64 & 76.73 & 1.002 & 0.646 &
& 51.60  & \best{84.25} & \best{84.14} & \best{0.567} & \best{0.728} \\

OMIB 
& \best{41.61}  & 81.07 & 81.08 & 0.834 & 0.711 &
& 51.04  & 83.95 & 83.89 & 0.579 & 0.727 \\

QMF 
& 38.54  & 80.15 & 80.16 & 0.876 & 0.695 &
& 48.52  & 82.65 & 82.41 & 0.606 & 0.707 \\

PML 
& 35.18  & 77.71 & 77.84 & 0.950 & 0.667 &
& 51.41  & 83.18 & 82.96 & 0.581 & 0.715 \\

HME 
& 26.86  & 66.56 & 66.33 & 1.266 & 0.464 &
& \best{51.93}  & 83.43 & 83.45 & 0.574 & 0.714 \\

\textbf{RCE (Ours)} 
& 40.29  & \best{81.07} & \best{80.62} & \best{0.820} & \best{0.723} &
& 50.89  & 82.27 & 82.14 & 0.584 & 0.705 \\

\midrule 
\rowcolor{offlinecolor} \multicolumn{12}{c}{\textit{\textbf{Salt-pepper Noise Intensity: 10.0}}} \\ 
\midrule

C-MIB 
& 31.09  & 67.94 & 67.99 & 1.059 & 0.579 &
& 46.59  & 73.34 & 71.85 & 0.694 & 0.558 \\

ITHP 
& 26.57  & 61.83 & 60.35 & 1.157 & 0.528 &
& 45.68  & 75.17 & \best{74.36} & 0.702 & 0.572 \\

KAN-MCP 
& 31.97  & 67.18 & 66.17 & 1.084 & \best{0.603} &
& 48.24  & 75.08 & 72.80 & 0.675 & 0.568 \\

OMIB 
& 28.61  & 65.34 & 65.10 & 1.096 & 0.538 &
& 46.87  & 75.11 & 73.48 & 0.688 & 0.546 \\

QMF 
& 32.85  & 65.04 & 64.01 & 1.055 & 0.583 &
& 46.22  & 74.64 & 71.87 & 0.698 & 0.530 \\

PML 
& 29.34  & 68.40 & 67.71 & 1.048 & 0.582 &
& 47.38  & 74.39 & 72.68 & 0.678 & 0.569 \\

HME 
& 31.97  & 67.48 & 67.63 & 1.045 & 0.575 &
& 46.20  & 74.56 & 72.58 & 0.690 & 0.546 \\

\textbf{RCE (Ours)} 
& \best{33.28}  & \best{73.44} & \best{72.35} & \best{1.019} & 0.586 &
& \best{48.26}  & \best{75.33} & 73.27 & \best{0.668} & \best{0.576} \\

\bottomrule[1.5pt]

\end{tabular}
}
\end{table}

\paragraph{Results under Gaussian noise.}
Table~\ref{tab:noise_gaussian} reports the comparison under different Gaussian noise intensities. Overall, RCE demonstrates strong robustness across both datasets, especially under moderate and severe Gaussian perturbations. When the noise intensity is 1.0, RCE achieves the best results on most metrics. On CMU-MOSI, RCE obtains the best $Acc7$, $Acc2$, $F1$, and $MAE$, while its $Corr$ remains very close to the best result. On CMU-MOSEI, RCE achieves the best $Acc7$, $MAE$, and $Corr$, and remains highly competitive on $Acc2$ and $F1$. These results indicate that when the perturbation is mild, RCE can effectively preserve useful sentiment cues while suppressing unreliable corrupted information.

As the Gaussian noise intensity increases to 5.0, all methods experience performance degradation, but RCE still maintains competitive performance. On CMU-MOSI, RCE achieves the best $Acc2$ and $F1$, showing that it preserves binary sentiment discrimination under stronger feature perturbation, although HME~\cite{HME2025} obtains better $Acc7$, $MAE$, and $Corr$ in this setting. On CMU-MOSEI, RCE achieves the best $Acc2$, $F1$, and $MAE$, demonstrating stable performance on the larger-scale dataset. Under the most severe Gaussian noise intensity of 10.0, RCE shows clearer advantages. On CMU-MOSI, it achieves the best performance across all five metrics, improving $Acc2$ and $F1$ notably compared with the strongest baselines. On CMU-MOSEI, RCE also obtains the best $Acc7$, $Acc2$, and $MAE$, while remaining close to the best $Corr$. These results suggest that the reliability-aware enhancement and multilevel fusion mechanisms help RCE maintain robust predictions when acoustic and visual observations are heavily corrupted.

\paragraph{Results under Salt-pepper noise.}
Table~\ref{tab:noise_salt} presents the results under Salt-pepper noise with different corruption intensities. Compared with Gaussian noise, Salt-pepper noise introduces more abrupt feature-level corruption by replacing valid acoustic and visual elements with extreme values from the same sample, making it a challenging test of robustness to outlier-like perturbations~\cite{QMF2023}. When the noise intensity is 1.0, RCE achieves the best $Acc7$, $Acc2$, $F1$, and $MAE$ on CMU-MOSI, and the best $Acc7$, $Acc2$, $F1$, and $Corr$ on CMU-MOSEI. This shows that RCE is effective under mild sparse corruption and can still extract reliable complementary information from noisy continuous features.

When the Salt-pepper noise intensity increases to 5.0, the performance of all methods decreases more noticeably. On CMU-MOSI, RCE achieves the best $Acc2$, $F1$, $MAE$, and $Corr$, and remains close to the best $Acc7$. This indicates that RCE is particularly effective at preserving overall sentiment discrimination and regression quality under medium-level outlier corruption. On CMU-MOSEI, however, KAN-MCP~\cite{KAN-MCP2025} and HME~\cite{HME2025} obtain better results on several metrics, while RCE becomes less competitive. This suggests that medium Salt-pepper corruption may introduce feature-level outliers that affect RCE's retrieved or fused representations on the larger dataset. Nevertheless, under the strongest Salt-pepper noise intensity of 10.0, RCE again shows robust overall performance. It achieves the best $Acc7$, $Acc2$, $F1$, and $MAE$ on CMU-MOSI, and the best $Acc7$, $Acc2$, $MAE$, and $Corr$ on CMU-MOSEI. These results demonstrate that RCE is generally resilient to severe outlier-style corruption. The advantage is especially clear under high noise intensity, where reliability-aware fusion can reduce the influence of unreliable modality information and better exploit stable cues from the remaining useful representations.


\subsection{Detailed Ablation Study}
\label{appendix:ablation_study}

\begin{table}[h]
\centering  
\setlength{\belowcaptionskip}{10pt}
\renewcommand\arraystretch{1.1}
\caption{Ablation details.}

\label{tab:ablation_details}
\resizebox{\textwidth}{!}{

\begin{tabular}{lccccccccccc}

\toprule[1.5pt]

\multirow{2}{*}{\textbf{Method}} 
& \multicolumn{5}{c}{\textbf{CMU-MOSI}} & & \multicolumn{5}{c}{\textbf{CMU-MOSEI}} \\
\cmidrule{2-6} \cmidrule{8-12}
& \textit{Acc7\%$\uparrow$} & \textit{Acc2\%$\uparrow$} & \textit{F1\%$\uparrow$} & \textit{MAE$\downarrow$} &\textit{Corr$\uparrow$} & & \textit{Acc7\%$\uparrow$} & \textit{Acc2\%$\uparrow$} & \textit{F1\%$\uparrow$} & \textit{MAE$\downarrow$} &\textit{Corr$\uparrow$} \\

\midrule

w/o AVIB
& 48.03 & 87.79 & 87.77 & 0.646 & 0.842 &
& 54.60 & 86.35 & 86.47 & 0.518 & 0.785 \\

w/o RME
& \best{50.36} & 88.09 & 88.02 & 0.610 & 0.857 &
& 54.10 & 87.04 & 87.04 & 0.519 & 0.794 \\

w/o MB
& 48.03 & 88.24 & 88.20 & 0.616 & 0.853 &
& 53.50 & 87.18 & 87.15 & 0.511 & 0.794 \\

w/o LP
& 49.34 & \second{88.70} & \second{88.62} & \second{0.603} & \second{0.859} &
& \best{56.06} & 86.60 & 86.67 & 0.505 & 0.795 \\

w/o CP
& 47.30 & 87.94 & 87.81 & 0.610 & 0.856 &
& 55.14 & 87.18 & 87.17 & 0.504 & \second{0.798} \\

w/o HMG
& 46.72 & 87.63 & 87.56 & 0.637 & 0.841 &
& 54.77 & 86.82 & 86.76 & 0.506 & 0.794 \\

w/o ES
& 45.69 & 87.94 & 87.82 & 0.654 & 0.842 &
& 55.33 & \best{87.51} & \second{87.42} & \best{0.502} & 0.797 \\

w/o RF
& 48.47 & 87.33 & 87.20 & 0.626 & 0.851 &
& \second{55.52} & \best{87.51} & \best{87.47} & 0.507 & 0.796 \\

RCE
& \second{50.22} & \best{89.31} & \best{89.26} & \best{0.589} & \best{0.861} &
& 55.44 & \second{87.29} & 87.33 & \second{0.503} & \best{0.799} \\

\bottomrule[1.5pt]

\end{tabular}
}
\end{table}

To further examine the contribution of each component in RCE, we report detailed ablation results in Table~\ref{tab:ablation_details}. Specifically, `w/o AVIB' removes the Adaptive Variational Information Bottleneck, where modality representations are directly used without vMF-based uncertainty estimation, latent sampling, and KL regularization. `w/o RME' removes the Reliability-aware Modality Enhancement module by directly setting the enhanced representation as the original modality representation. `w/o MB' keeps the enhancement mechanism but disables the memory bank, using only mini-batch samples for neighbor retrieval. `w/o LP' removes the label-distance penalty in neighbor weighting, while `w/o CP' removes the confidence promotion term based on $\kappa$. `w/o HMG' removes the hyper-modality generation module and directly uses modality-level representations for fusion. `w/o ES' removes the shared sample-level representation, and `w/o RF' disables reliability-aware fusion by replacing $\kappa$-modulated fusion with vanilla attention weights.

Overall, the complete RCE achieves the best or competitive results on most metrics, especially on CMU-MOSI, where it obtains the best $Acc2$, $F1$, $MAE$, and $Corr$. This indicates that the proposed components are complementary and jointly improve both classification accuracy and regression quality. On CMU-MOSI, removing ES leads to the largest degradation in $Acc7$ and $MAE$, with $Acc7$ dropping from 50.22 to 45.69 and $MAE$ increasing from 0.589 to 0.654, showing that the shared sample-level representation is important for capturing global multimodal context. Removing HMG also causes a clear performance drop, especially in $Acc7$ and $MAE$, indicating that hyper-modality generation contributes to robust cross-modal representation learning. In addition, removing AVIB consistently weakens all metrics on CMU-MOSI, demonstrating that vMF-based uncertainty estimation and adaptive compression help reduce noisy and redundant modality information. The memory bank also plays an important role: compared with RCE, `w/o MB' obtains lower $Acc7$, $MAE$, and $Corr$, suggesting that historical reliable samples provide more stable neighborhood information than mini-batch-only retrieval.

For the internal design of RME, the effects of label penalty and confidence promotion show different tendencies. On CMU-MOSI, `w/o LP' achieves relatively strong classification results but is still inferior to RCE in $Acc2$, $F1$, $MAE$, and $Corr$, suggesting that label consistency is beneficial for stable prediction. In contrast, removing CP causes a larger decline in $Acc7$, indicating that confidence-aware neighbor weighting helps suppress unreliable enhancement samples. On CMU-MOSEI, several ablated variants obtain slightly better results on individual metrics, such as `w/o LP' on $Acc7$ and `w/o ES' or `w/o RF' on $Acc2$/$F1$. This may be because CMU-MOSEI contains more training samples and provides more stable multimodal statistics, making some auxiliary constraints less critical for coarse-grained classification. Nevertheless, the complete RCE still achieves the best $Corr$ and nearly the best $MAE$, showing better overall regression consistency and prediction calibration. In particular, although `w/o RF' slightly improves $Acc2$ and $F1$ on CMU-MOSEI, it performs worse than RCE in $MAE$ and $Corr$, indicating that reliability-aware fusion is more helpful for fine-grained sentiment intensity estimation. Overall, these results show that AVIB, RME with memory-based retrieval, HMG, ES, and RF contribute from different aspects, and their combination yields the most balanced performance across datasets and evaluation metrics.


\subsection{Negative Enhancement Analysis}
\label{appendix:analysis_negative_enhancement}

\begin{table*}[h]
\centering
\caption{Detailed training-stage negative enhancement statistics on CMU-MOSI and CMU-MOSEI.}
\label{tab:negative_enhancement_detail}
\resizebox{\linewidth}{!}{
\begin{tabular}{cccccc}
\toprule[1.5pt]
\textbf{Dataset} & \textbf{Method} & \textbf{Modality} & 
\textbf{Samples} & \textbf{Negative Enhancement Samples} & \textbf{Negative Enhancement Rate} \\

\midrule
\multirow{6}{*}{CMU-MOSI}
& \multirow{3}{*}{HME}
& Text  & 61,400 & 28,420 & 46.29\% \\
& & Audio & 61,400 & 28,797 & 46.90\% \\
& & Vision & 61,400 & 28,797 & 46.90\% \\
\cmidrule(lr){2-6}
& \multirow{3}{*}{RCE}
& Text  & 61,400 & 1,595 & \best{2.60\%} \\
& & Audio & 61,400 & 3,621 & \best{5.90\%} \\
& & Vision & 61,400 & 3,976 & \best{6.48\%} \\
\midrule
\multirow{6}{*}{CMU-MOSEI}
& \multirow{3}{*}{HME}
& Text  & 127,350 & 48,904 & 38.40\% \\
& & Audio & 127,350 & 49,500 & 38.87\% \\
& & Vision & 127,350 & 49,500 & 38.87\% \\
\cmidrule(lr){2-6}
& \multirow{3}{*}{RCE}
& Text  & 127,350 & 13,458 & \best{10.57\%} \\
& & Audio & 127,350 & 39,798 & \best{31.25\%} \\
& & Vision & 127,350 & 33,218 & \best{26.08\%} \\
\bottomrule[1.5pt]
\end{tabular}
}
\end{table*}

We further analyze the negative enhancement phenomenon during training. For each sample, we compare the sentiment direction of its ground-truth label with the sentiment direction of the samples used for enhancement. Specifically, for HME, the retrieved samples whose similarities exceed the threshold are averaged to obtain the enhancement source. For RCE, we use the final weighted neighbors after top-$k$ retrieval, label-distance penalty, and reliability-aware weighting. If the weighted average label of the enhancement sources has the opposite sentiment polarity to the current sample, this case is counted as negative enhancement.

The reported rate is computed as the number of negative enhancement cases divided by the number of valid enhanced samples in the training stage. In our logs, CMU-MOSI contains 61,400 valid training enhancement cases over all epochs, corresponding to 1,228 valid training samples per epoch, while CMU-MOSEI contains 127,350 valid training enhancement cases, corresponding to 12,735 valid training samples per epoch. Neutral samples are not counted since their sentiment polarity is ambiguous.

As shown in Table~\ref{tab:negative_enhancement_rate}, HME suffers from severe negative enhancement. On CMU-MOSI, nearly 46\% of training samples are negatively enhanced across all modalities. In contrast, RCE reduces the negative enhancement rates to 2.60\%, 5.90\%, and 6.48\% for text, audio, and vision, respectively. This demonstrates that the proposed label-aware and reliability-aware enhancement mechanism effectively suppresses sentiment-conflicting information from neighboring samples.

On CMU-MOSEI, RCE also consistently reduces negative enhancement compared with HME. The reduction is most significant in the text modality, from 38.40\% to 10.57\%. However, the decrease for the audio modality is relatively smaller. This may be because acoustic cues in CMU-MOSEI are more ambiguous and diverse: samples with similar acoustic patterns may still express different sentiment polarities. As a result, even after reliability-aware weighting, some sentiment-conflicting audio neighbors can still receive non-negligible weights, leading to a higher remaining negative enhancement rate.


\begin{figure*}[h]
    \centering

    \begin{subfigure}[h]{0.32\linewidth}
        \centering
        \includegraphics[width=\linewidth]{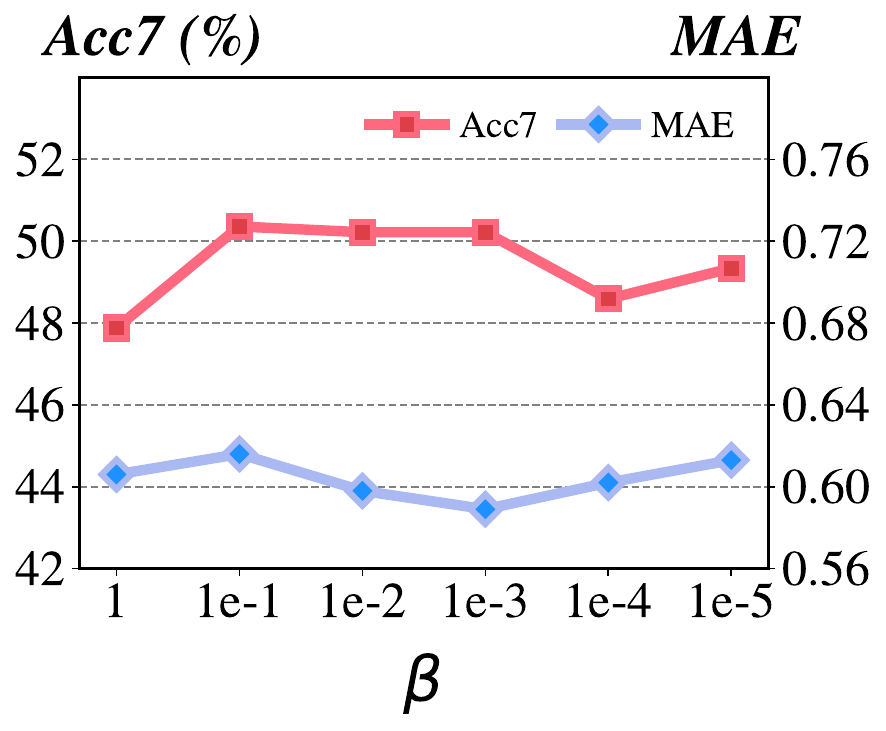}
        \caption{AVIB Trade-off Coefficient $\beta$.}
        \label{fig:mosi_alpha}
    \end{subfigure}
    \begin{subfigure}[h]{0.32\linewidth}
        \centering
        \includegraphics[width=\linewidth]{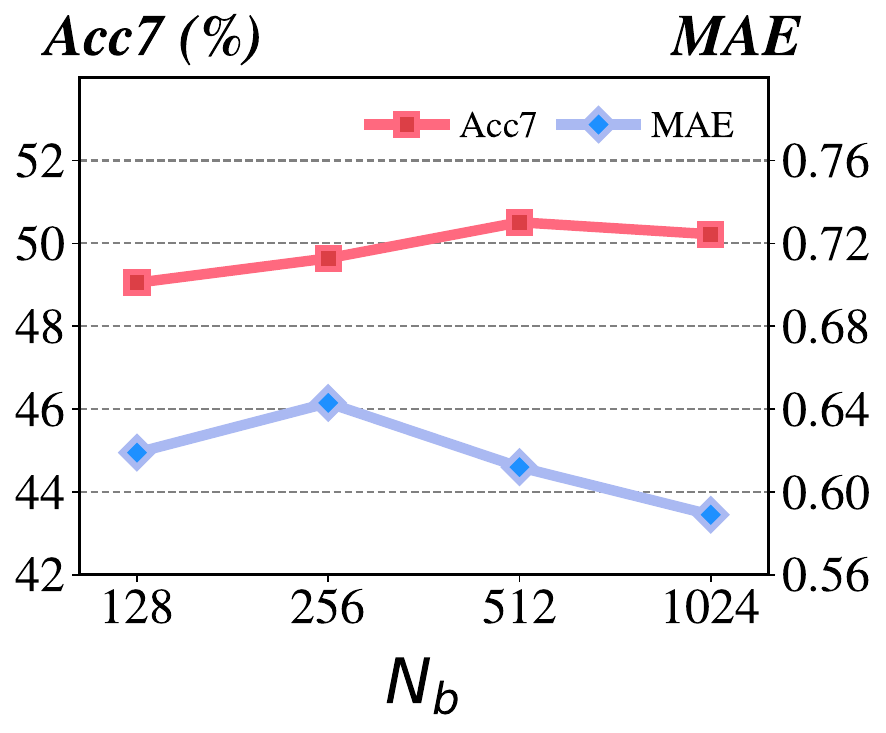}
        \caption{Memory Bank Capacity $N_b$.}
        \label{fig:mosi_lambda}
    \end{subfigure}

    \caption{Parameter sensitivity analysis of $\beta$ and $N_b$ on the CMU-MOSI dataset.}
    \label{fig:mosi_parameter_analysis_1}
\end{figure*}

\begin{figure*}[h]
    \centering

    \begin{subfigure}[h]{0.32\linewidth}
        \centering
        \includegraphics[width=\linewidth]{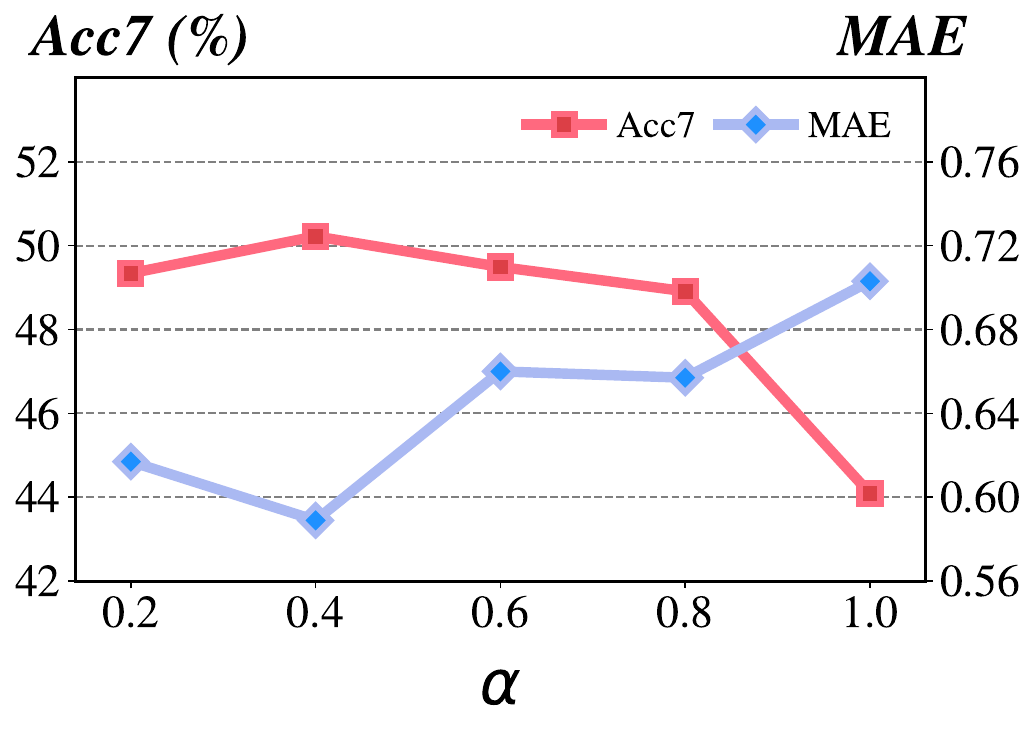}
        \caption{Label Penalty Weight $\alpha$.}
        \label{fig:mosi_alpha}
    \end{subfigure}
    \hfill
    \begin{subfigure}[h]{0.32\linewidth}
        \centering
        \includegraphics[width=\linewidth]{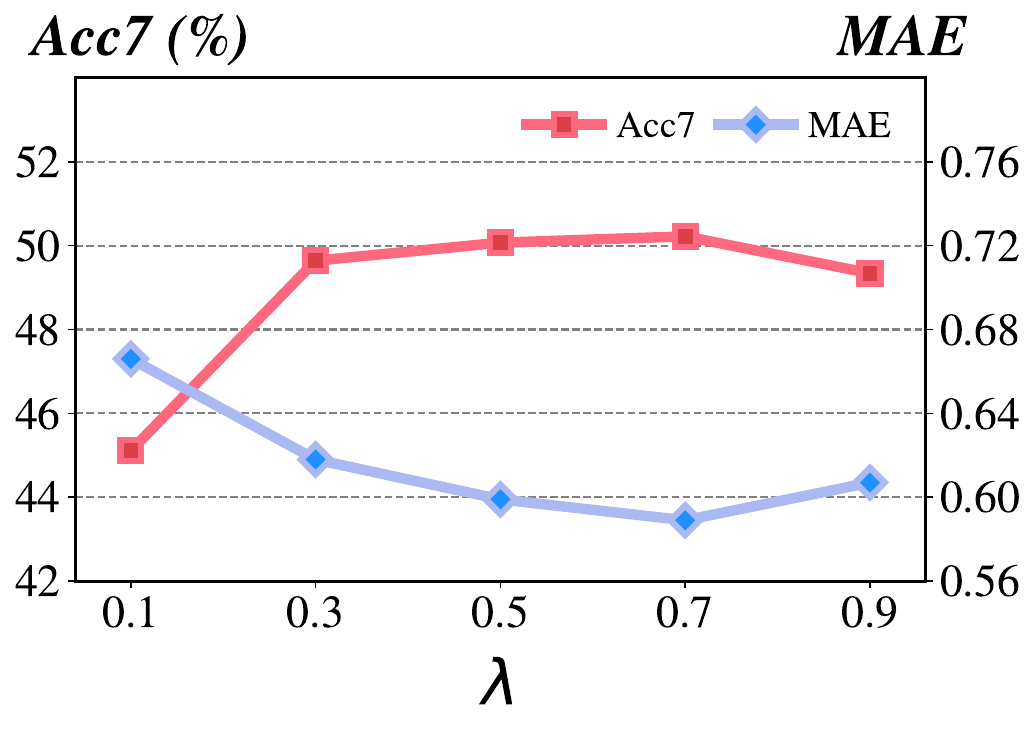}
        \caption{ Loss Balancing Weight $\lambda$.}
        \label{fig:mosi_beta}
    \end{subfigure}
    \hfill
    \begin{subfigure}[h]{0.32\linewidth}
        \centering
        \includegraphics[width=\linewidth]{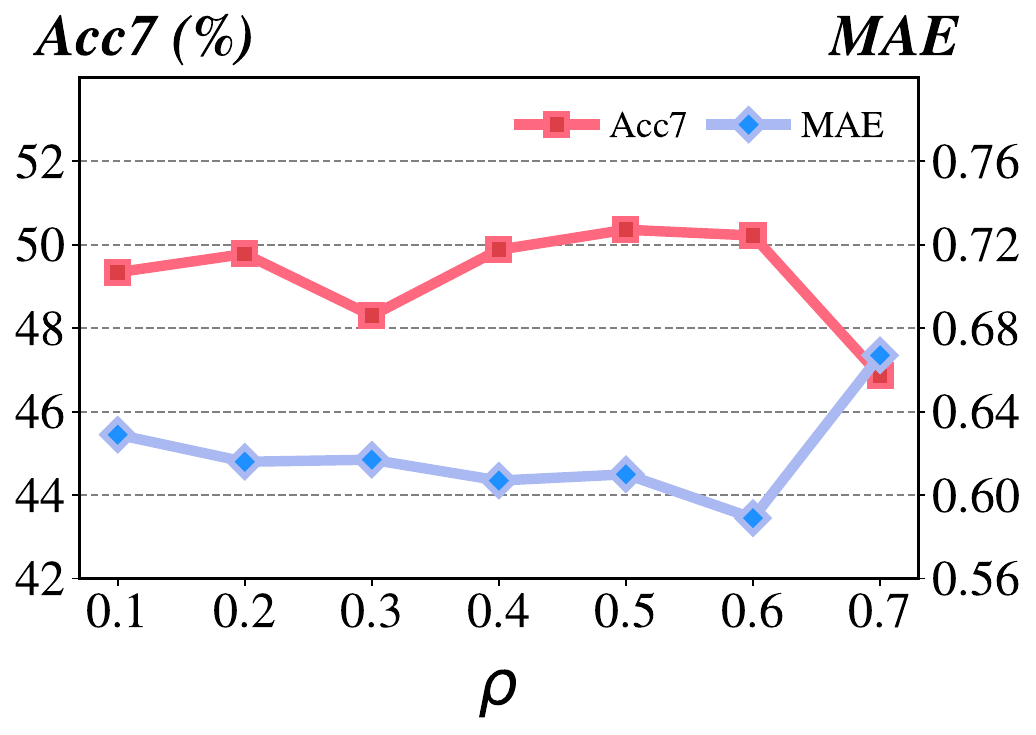}
        \caption{Top-$k$ Ratio $\rho$.}
        \label{fig:mosi_lambda}
    \end{subfigure}

    \caption{Parameter sensitivity analysis of $\alpha$, $\lambda$, and $\rho$ on the CMU-MOSI dataset.}
    \label{fig:mosi_parameter_analysis_2}
\end{figure*}

\subsection{Parameter Sensitivity Analysis}
\label{appendix:parameter_sensitivity}

\begin{figure*}[h]
    \centering

    \begin{subfigure}[h]{0.32\linewidth}
        \centering
        \includegraphics[width=\linewidth]{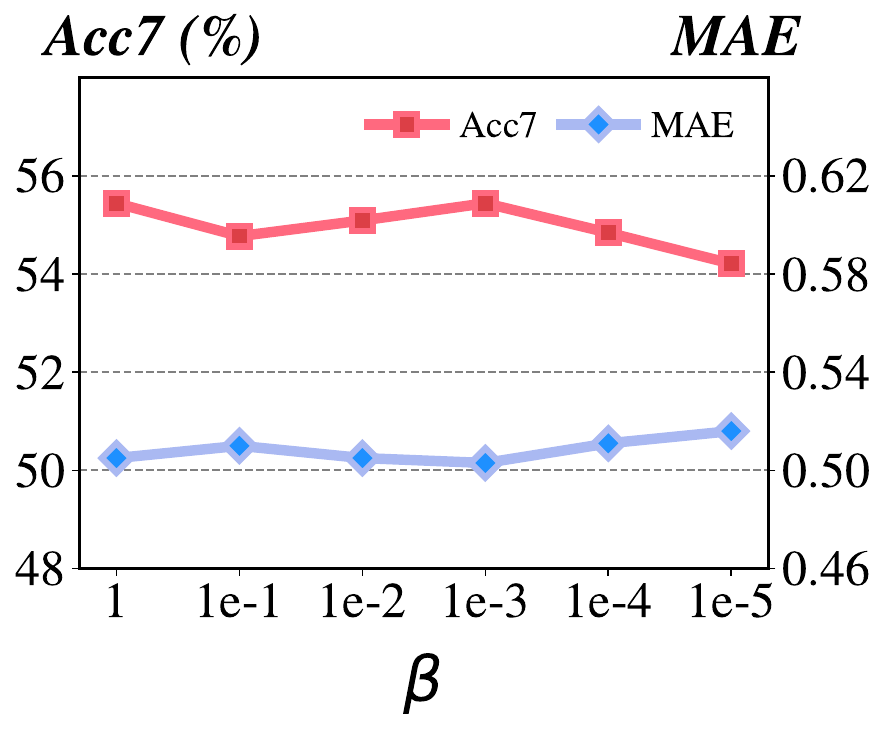}
        \caption{AVIB Trade-off Coefficient $\beta$.}
        \label{fig:mosei_beta}
    \end{subfigure}
    \begin{subfigure}[h]{0.32\linewidth}
        \centering
        \includegraphics[width=\linewidth]{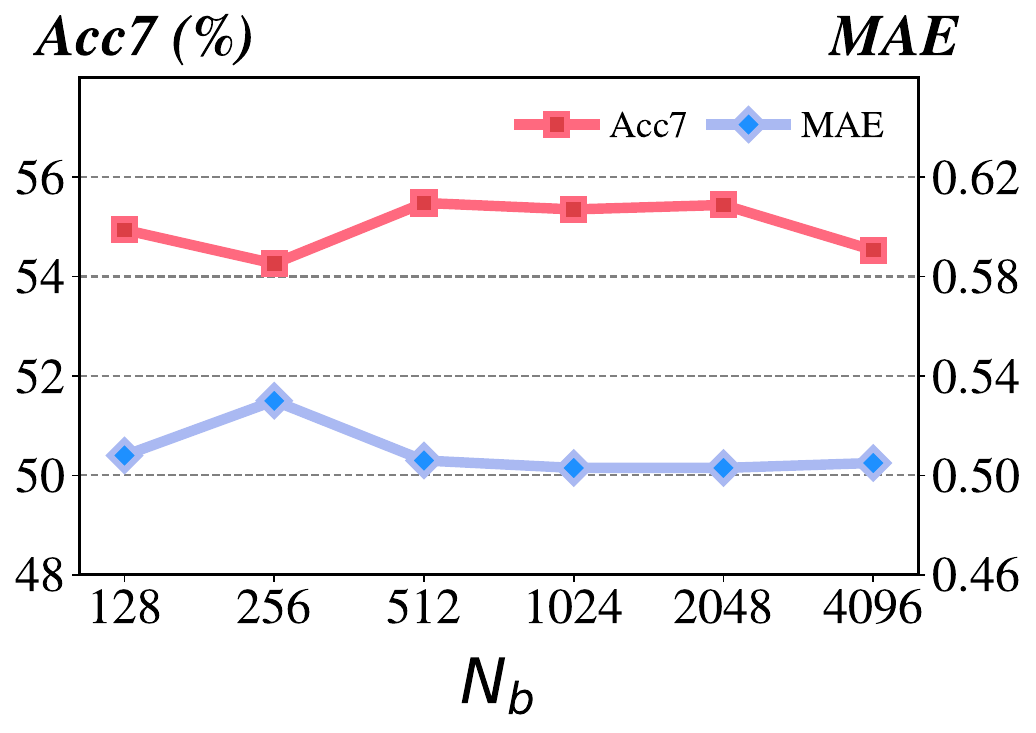}
        \caption{Memory Bank Capacity $N_b$.}
        \label{fig:mosei_mb}
    \end{subfigure}

    \caption{Parameter sensitivity analysis of $\beta$ and $N_b$ on the CMU-MOSEI dataset.}
    \label{fig:mosei_parameter_analysis_1}
\end{figure*}

\begin{figure*}[h]
    \centering

    \begin{subfigure}[h]{0.32\linewidth}
        \centering
        \includegraphics[width=\linewidth]{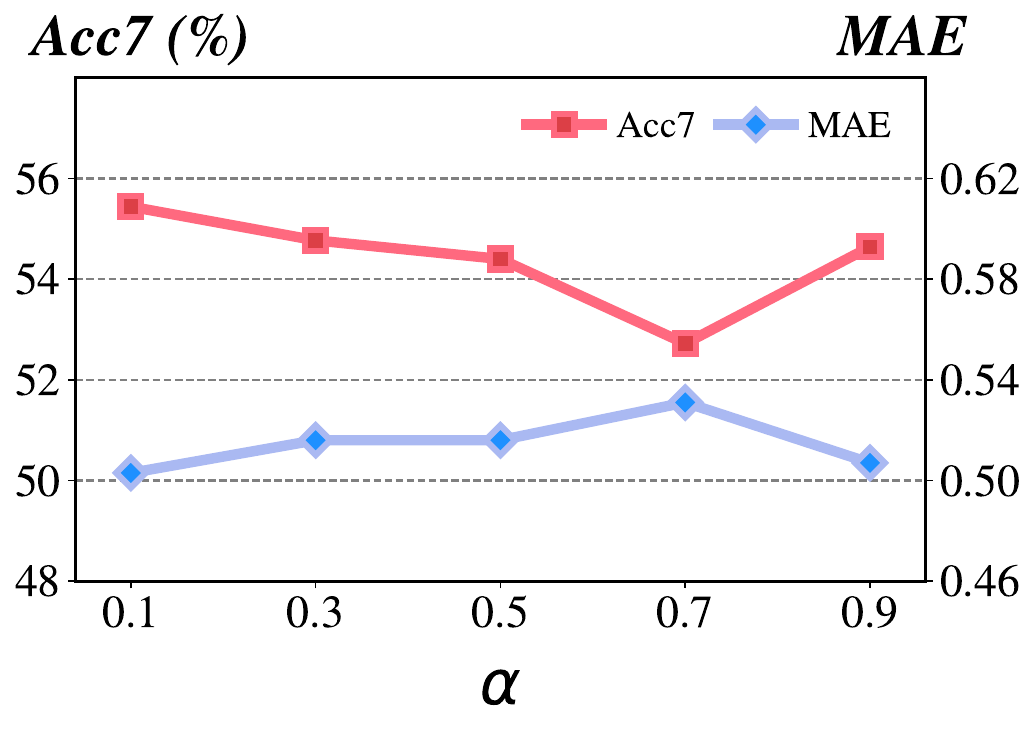}
        \caption{Label Penalty Weight $\alpha$.}
        \label{fig:mosei_alpha}
    \end{subfigure}
    \hfill
    \begin{subfigure}[h]{0.32\linewidth}
        \centering
        \includegraphics[width=\linewidth]{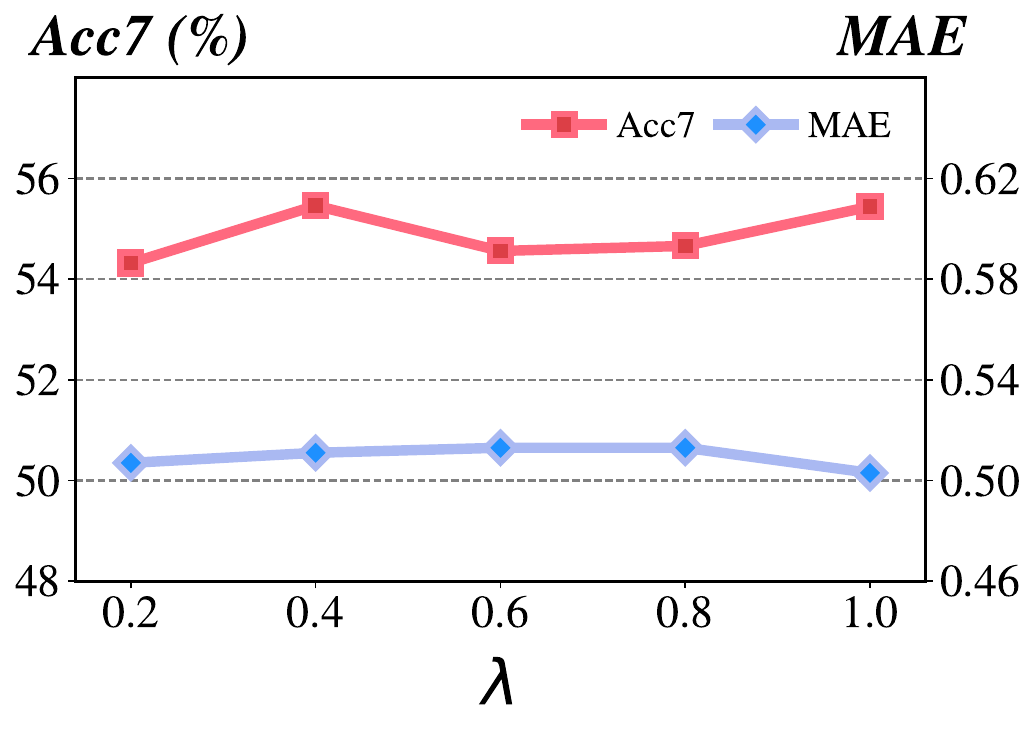}
        \caption{Loss Balancing Weight $\lambda$.}
        \label{fig:mosei_lambda}
    \end{subfigure}
    \hfill
    \begin{subfigure}[h]{0.32\linewidth}
        \centering
        \includegraphics[width=\linewidth]{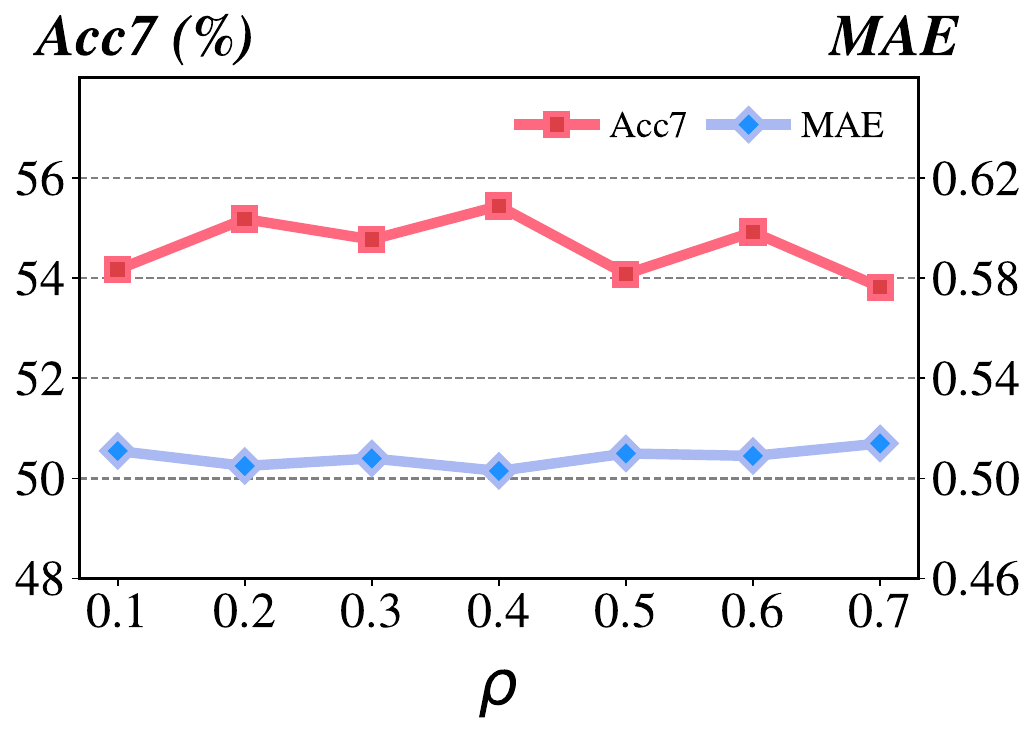}
        \caption{Top-$k$ Ratio $\rho$.}
        \label{fig:mosei_topk}
    \end{subfigure}

    \caption{Parameter sensitivity analysis of $\alpha$, $\lambda$, and $\rho$ on the CMU-MOSEI dataset.}
    \label{fig:mosei_parameter_analysis_2}
\end{figure*}

We analyze the sensitivity of RCE to five important hyperparameters, including the AVIB trade-off coefficient $\beta$, the memory bank capacity $N_b$, the label penalty weight $\alpha$, the loss balancing weight $\lambda$, and the top-$k$ ratio $\rho$. The results on CMU-MOSI and CMU-MOSEI are shown in Fig.~\ref{fig:mosi_parameter_analysis_1}, Fig.~\ref{fig:mosi_parameter_analysis_2}, Fig.~\ref{fig:mosei_parameter_analysis_1}, and Fig.~\ref{fig:mosei_parameter_analysis_2}.

\textbf{(1) AVIB trade-off coefficient $\beta$.}
The coefficient $\beta$ controls the strength of the information bottleneck regularization in AVIB. On CMU-MOSI, RCE obtains competitive performance under a wide range of $\beta$ values, with the best $Acc7$ achieved around $\beta=1e$-$1$ and the best $MAE$ obtained at $\beta=1e$-$3$. On CMU-MOSEI, the model is also relatively stable, and $\beta=1e$-$3$ achieves both strong $Acc7$ and the lowest $MAE$. These results indicate that a moderate bottleneck constraint helps remove noisy or redundant information, while overly large or overly small values may weaken either task-relevant information preservation or regularization effectiveness. Therefore, we adopt $\beta=1e$-$3$ as the default setting.

\textbf{(2) Memory bank capacity $N_b$.}
The memory bank capacity determines the number of historical samples available for reliability-aware modality enhancement. On both datasets, very small memory banks provide limited neighborhood information and lead to inferior performance. As $N_b$ increases, the performance generally improves, showing that a larger candidate pool can provide more stable and informative enhancement samples. On CMU-MOSI, $N_b=512$ achieves the best $Acc7$, while $N_b=1024$ obtains the lowest $MAE$. On CMU-MOSEI, $N_b=512$ also achieves the best $Acc7$, and $N_b=1024$ or $2048$ gives the best $MAE$. This suggests that a moderate-to-large memory bank is beneficial, but continuously increasing the capacity does not always bring further gains.

\textbf{(3) Label penalty weight $\alpha$.}
The label penalty weight $\alpha$ controls how strongly label discrepancy is penalized during neighbor weighting. On CMU-MOSI, the model performs best around $\alpha=0.4$, where both $Acc7$ and $MAE$ achieve the best results. When $\alpha$ becomes too large, performance drops clearly, suggesting that overly strong label constraints may suppress useful complementary samples. On CMU-MOSEI, the performance is relatively stable for most $\alpha$ values, although $\alpha=0.1$ gives the best result. This may be because CMU-MOSEI contains more samples and more diverse sentiment expressions, making a mild label penalty sufficient for filtering conflicting neighbors.

\textbf{(4) Loss balancing weight $\lambda$.}
The weight $\lambda$ balances the main prediction loss and the AVIB-related auxiliary objective. On CMU-MOSI, RCE achieves the best performance when $\lambda=0.7$, indicating that a proper contribution from AVIB is important for robust sentiment prediction. On CMU-MOSEI, the model remains stable across different $\lambda$ values, with the best $Acc7$ around $\lambda=0.4$ and the lowest $MAE$ around $\lambda=1.0$. Overall, moderate values of $\lambda$ provide a good balance between prediction accuracy and representation regularization.

\textbf{(5) Top-$k$ ratio $\rho$.}
The top-$k$ ratio $\rho$ controls the proportion of retrieved neighbors used for modality enhancement. If $\rho$ is too small, the model may not obtain sufficient complementary information. If $\rho$ is too large, less relevant or sentiment-conflicting samples may be introduced. On CMU-MOSI, the best $Acc7$ is obtained around $\rho=0.5$, while the lowest $MAE$ is achieved at $\rho=0.6$. On CMU-MOSEI, $\rho=0.4$ achieves the best overall performance. These results show that selecting a moderate number of reliable neighbors is more effective than using either too few or too many candidates.

Overall, RCE is robust to these hyperparameters on both datasets. Although different datasets prefer slightly different values, the performance remains stable within moderate ranges. This indicates that the proposed reliability-aware enhancement and multilevel fusion mechanisms are not overly sensitive to hyperparameter choices.


\subsection{Testing Stability Analysis}
\label{appendix:testing_stability}

Table~\ref{tab:testing_stability} reports the testing stability of RCE on the CMU-MOSI and CMU-MOSEI datasets. For each dataset, we run RCE with five different random seeds and compute the mean performance together with the 95\% confidence interval for each metric. Specifically, the confidence interval is calculated as $\bar{x} \pm t_{0.975,4} \frac{s}{\sqrt{5}}$, where $\bar{x}$ and $s$ denote the mean and standard deviation over the five runs, respectively. The narrow intervals on both CMU-MOSI and CMU-MOSEI show that RCE has small performance variation across different random initializations, indicating stable and reproducible results.

\begin{table}[h]
\centering
\caption{Testing stability with five random seeds under the complete-modality setting (95\% confidence intervals).}
\label{tab:testing_stability}
\resizebox{0.85\linewidth}{!}{
\begin{tabular}{cccccc}
\toprule[1.5pt]
\textbf{Dataset} & \textbf{$Acc7\%$} & \textbf{$Acc2\%$} & \textbf{$F1\%$} & \textbf{$MAE$} & \textbf{$Corr$} \\
\midrule
CMU-MOSI
& $49.81 \pm 0.56$
& $88.98 \pm 0.97$
& $88.93 \pm 0.97$
& $0.598 \pm 0.009$
& $0.859 \pm 0.004$ \\
CMU-MOSEI
& $55.14 \pm 0.37$
& $87.25 \pm 0.35$
& $87.26 \pm 0.31$
& $0.510 \pm 0.006$
& $0.792 \pm 0.006$ \\
\bottomrule[1.5pt]
\end{tabular}
}
\end{table}


\subsection{Model Complexity Analysis}
\label{appendix:model_complexity_analysis}
We compare the computational complexity of RCE with state-of-the-art methods in terms of FLOPs, memory usage, parameter count, and training time. As shown in Table~\ref{tab:complexity}, RCE introduces moderate additional cost due to the adaptive variational information bottleneck, memory bank-based reliability-aware modality enhancement, and multilevel fusion modules. Specifically, RCE requires 12.660G FLOPs and 9.49GB memory, which are comparable to most baselines and only slightly higher than HME~\cite{HME2025}. Although RCE has the largest parameter count, its training time is 8.54s per epoch, lower than both HME~\cite{HME2025} and KAN-MCP~\cite{KAN-MCP2025}. This indicates that the proposed enhancement and fusion mechanisms do not incur prohibitive overhead. Overall, RCE improves robustness and prediction performance while maintaining acceptable computational complexity.

\begin{table}[h]
\centering
\caption{Comparison of model complexity on the CMU-MOSI dataset with batch size 48.}
\label{tab:complexity}
\resizebox{0.8\textwidth}{!}{
\begin{tabular}{lcrcc}
\toprule[1.5pt]
\textbf{Method }
& \textbf{FLOPs} & \textbf{Memory} & \textbf{Parameters} & \textbf{Per-Epoch Training Time} \\
\midrule

C-MIB~\cite{MIB2023} 
& 11.773 G & 8.48 GB & 190.93 M & 2.93 s \\ 

ITHP~\cite{ITHP2024} 
& 11.523 G & 8.36 GB & 184.88 M & 3.43 s \\ 

KAN-MCP~\cite{KAN-MCP2025} 
& 11.501 G & 13.36 GB & 184.76 M & 10.62 s \\ 

OMIB~\cite{OMIB2025} 
& 11.615 G & 8.10 GB & 189.05 M & 3.70 s \\ 

QMF~\cite{QMF2023} 
& 12.529 G & 8.41 GB & 196.47 M & 4.09 s \\ 

PML~\cite{PML2025} 
& 12.144 G & 8.42 GB & 188.27 M & 3.99 s \\ 

HME~\cite{HME2025} 
& 12.407 G & 9.02 GB & 228.53 M & 10.30 s \\ 

\textbf{RCE (Ours)}
& 12.660 G & 9.49 GB & 251.77 M & 8.54 s \\ 

\toprule[1.5pt]
\end{tabular}  
}
\end{table}


\subsection{Theoretical Complexity Analysis}
\label{appendix:theoretical_complexity_analysis}
We further analyze the computational complexity of the key components in RCE. Let $B$ denote the batch size, $d$ the hidden dimension, $M$ the number of modalities, $N_b$ the memory bank capacity, and $\rho$ the top-$k$ retrieval ratio. The AVIB module mainly consists of vMF parameter estimation, latent sampling, and auxiliary prediction, with a complexity of $O(MBd^2)$. The RME module retrieves reliable neighbors from the modality-specific memory bank. Its similarity computation requires $O(MBN_bd)$, and weighted aggregation over selected neighbors requires $O(MB\rho N_b d)$. The hyper-modality generation and multilevel fusion modules are based on lightweight latent attention and fusion operations, with an approximate complexity of $O(MBd^2)$ since the number of latent prompts and fusion tokens is small and fixed. Therefore, the overall additional complexity of RCE is mainly dominated by memory-bank retrieval, i.e., $O(MBN_bd)$, while other components scale linearly with respect to the batch size and number of modalities. Since $M$ and $N_b$ are fixed hyperparameters in practice, RCE remains computationally tractable. This is also consistent with Table~\ref{tab:complexity}, where RCE achieves stronger robustness with only moderate additional cost compared to HME and other missing-modality baselines.


\clearpage



\end{document}